\documentclass[twoside,11pt]{article}
\usepackage{jair, theapa, rawfonts}

\ShortHeadings{Creative Problem Solving in Intelligent Agents: A Framework and Survey}
{Gizzi, Nair, Sinapov \& Chernova}
\firstpageno{1}

\usepackage[utf8]{inputenc} 
\usepackage[T1]{fontenc}    
\usepackage{url}            
\usepackage{booktabs}       
\usepackage{amsfonts}       
\usepackage{nicefrac}       
\usepackage{microtype}      
\usepackage{color}
\usepackage{graphicx}
\usepackage{subcaption}
\usepackage{amsmath}
\usepackage{amssymb}
\usepackage{tabularx}
\usepackage{csquotes}
\usepackage{tabu}
\usepackage{comment}
\usepackage{float}
\usepackage{supertabular}
\usepackage{longtable}
\usepackage[toc,page]{appendix}
\usepackage[normalem]{ulem}
\useunder{\uline}{\ul}{}

\newtheorem{definition}{Definition}
\graphicspath{{imgs/}}

\newcommand{\laks}[1]{{\textcolor[rgb]{0.0,0.0,1.0}{LN: {#1}}}}

\newcommand{\smallsec}[1]{\vspace{1.5mm} \noindent  \textbf{\small \textsc {#1} \\ }}

\newcommand{\boldsec}[1]{\vspace{1.5mm} \noindent  \textbf{\small \textsc {#1}:}}

\begin{document}

\title{Creative Problem Solving in Artificially Intelligent Agents: \\A Survey and Framework}
\date{}

\author{\name Evana Gizzi\thanks{Both authors contributed equally} \email evana.gizzi@tufts.edu \\
       \addr Department of Computer Science, Tufts University,\\
       Medford, MA 02155
       \AND
       \name Lakshmi Nair$^{*}$ \email lnair3@gatech.edu \\
       \addr College of Computing, Georgia Tech, \\
       Atlanta, GA 30332 
       \AND
       \name Sonia Chernova \email chernova@gatech.edu \\
       \addr College of Computing, Georgia Tech, \\
       Atlanta, GA 30332
       \AND
       \name Jivko Sinapov \email jivko.sinapov@tufts.edu \\
       \addr Department of Computer Science, Tufts University,\\
       Medford, MA 02155}



\maketitle

\begin{abstract}
Creative Problem Solving (CPS) is a sub-area within Artificial Intelligence (AI) that focuses on methods for solving off-nominal, or anomalous problems in autonomous systems. Despite many advancements in planning and learning, resolving novel problems or adapting existing knowledge to a new context, especially in cases where the environment may change in unpredictable ways post deployment, remains a limiting factor in the safe and useful integration of intelligent systems. The emergence of increasingly autonomous systems dictates the necessity for AI agents to deal with environmental uncertainty through creativity. To stimulate further research in CPS, we present a definition and a framework of CPS, which we adopt to categorize existing AI methods in this field. Our framework consists of four main components of a CPS problem, namely, 1) problem formulation, 2) knowledge representation, 3) method of knowledge manipulation, and 4) method of evaluation. We conclude our survey with open research questions, and suggested directions for the future.
\end{abstract}



\section{Introduction}
Creativity is often described as a hallmark of sophisticated intelligence. The Oxford English Dictionary defines ``creativity'' as ``\textit{Inventive, imaginative; of, relating to, displaying, using, or involving imagination or original ideas as well as routine skill or intellect, esp. in literature or art}'' \cite{dictionary1989oxford}. Despite our familiarity with the notion of creativity, understanding and implementing creativity in artificially intelligent systems continues to be a challenge. Computational Creativity (CC) is an active area of research that seeks to develop computational methods that are capable of generating a creative output, reminiscent of the creative processes in humans. The CC research community includes a diverse body of researchers, spanning the fields of psychology, neuroscience, philosophy, and computer science. The goal of CC research, as described by the Association for Computational Creativity\footnote{The Association of Computational Creativity is a nonprofit organization dedicated to the advancement of CC, and organizing body of the International Conference on Computational Creativity (ICCC)}, is \textit{``[to gain] the ability to model, simulate or replicate creativity using a computer, to achieve one of several ends, including the construction of a program or computer capable of human-level creativity, to better understand human creativity and to formulate an algorithmic perspective on creative behavior in humans, and to design programs that can enhance human creativity without necessarily being creative themselves''} \cite{CCdefintion}. While there has been extensive work in the area of Computational Creativity in Artificial Intelligence (AI), these works are primarily focused on the generation of creative artifacts, e.g., paintings, poems etc. In contrast, there is very limited focus on creativity that is specifically task-oriented, i.e., creativity in problem solving.

\begin{figure}
\centering
\begin{subfigure}{.5\textwidth}
  \centering
  \includegraphics[width=0.72\linewidth]{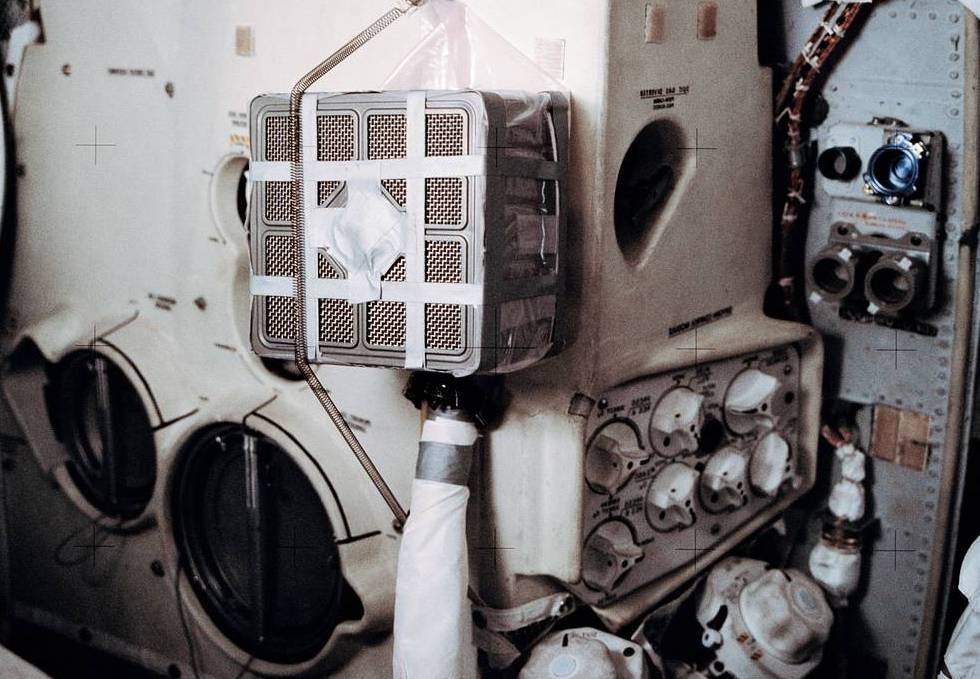}
  \label{fig:sub1}
\end{subfigure}%
\begin{subfigure}{.5\textwidth}
  \centering
  \includegraphics[width=0.82\linewidth]{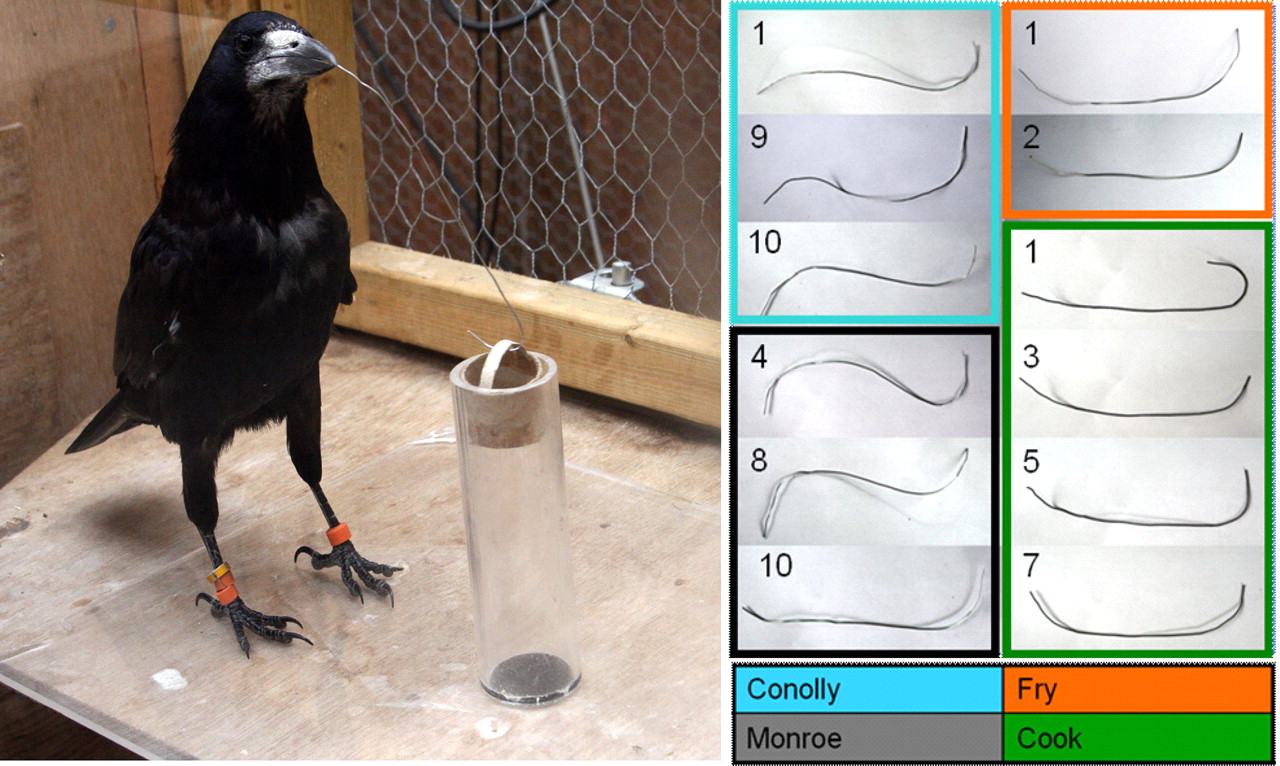}
  \label{fig:sub2}
\end{subfigure}
\caption{Examples of CPS in human and non-human species: The jury-rigged filter constructed by the astronauts on Apollo 13 (left, image credit: NASA). Rook extracting a bucket by bending a piece of wire to make a hook (right) \cite{bird2009insightful}.}
\label{fig:CPS_examples}
\end{figure}

Creative problem solving (CPS) focuses on using creative processes in the context of problem solving. Both human and non-human species have been shown to creatively solve problems \cite{boesch1990tool,baker2019emergent} (Figure \ref{fig:CPS_examples}), e.g., crows have been shown to spontaneously modify tools by shaping hooks out of wire, and using the modified tools in the correct sequence of actions required to retrieve food \cite{bird2009insightful}. Various models of CPS in humans have also been proposed by researchers over the years \cite{isaksen2004celebrating}, beginning as early as 1952 with the work of Alex Osborn, who presented a comprehensive description of a seven-stage CPS process \cite{osborn1952wake}. Prior work by Mumford et al. \cite{mumford1991process,mumford1997process} has also investigated CPS processes in humans by conducting human participant studies and evaluations. They define creative problems as problems that arise in ill-defined situations, thus eliciting creativity in humans. Further, their work highlights core cognitive processes involved in CPS, including problem construction, information encoding, idea evaluation, implementation, and monitoring. While these works primarily explore CPS in humans and other animals, there is very limited work focusing on CPS in artificial agents. 

\textbf{What makes CPS important for AI?} Numerous real-world examples demonstrate the practical importance of CPS, particularly when dealing with crises or time-constrained scenarios. For instance, in the Apollo 13 incident of 1970, astronauts on board the spacecraft creatively constructed a makeshift CO$_2$ filter using unconventional materials, enabling them to safely return home \cite{cass2005apollo} (See left, Figure \ref{fig:CPS_examples}). More recently, makeshift ventilators built using low-cost 3D printed parts and off-the-shelf items such as manual resuscitator PVC bags and motors, have been used to combat the widespread equipment shortages during COVID-19 \cite{ApolloBVM,turner2020thinking}. However, similar skills are currently beyond the scope of AI. Developing artificial agents with similar capabilities, can greatly improve the resourcefulness and adaptability of existing AI systems. These capabilities will be especially useful for robots that explore, as well as work in space, underwater, remote locations on land, and disaster sites, where the robots are highly likely to face unprecedented circumstances, requiring them to adapt.  \cite{atkeson2018happened}.


In this survey, we describe how CPS can leverage concepts from CC and planning/learning in AI to improve the adaptability of existing AI systems to novel scenarios. Similar surveys have previously focused specifically on CC by reviewing interdisciplinary work in CC along with evaluation techniques for CC systems (\cite{jordanous2013evaluating}, \cite{lamb2018evaluating}). Rowe et al. \cite{rowe1993creativity} surveyed CC works explicitly in an AI context, in which they suggest five key aspects of CC systems. These include a) flexibility of knowledge representations, b) tolerance to ambiguity in the knowledge representations, c) avoiding functional fixity, d) assessing the usefulness of the creative output, and e) the capacity to elaborate on the creative output to find out their consequences. The special issue journal on \textit{``Problem-solving, Creativity and Spatial Reasoning''} by Falomir et al. \cite{falomir2019special} compiled selected works in existing CPS research for a multi-disciplinary perspective on problem solving in CC, focused specifically on highlighting the synergies between the traditionally separate research areas. In contrast to prior surveys, to the best of our knowledge, this is the first survey that is specifically focused on creative problem solving in AI, leveraging the literature from \textit{both} CC and AI. Our goal in this survey is to contribute a taxonomy of research in CPS, and to provide organization and clarity into what has already been achieved as well as open research questions. We believe that a comprehensive discussion of CPS, combining CC and AI principles, is vital for encouraging future work in this area. 

Our paper is organized as follows. We begin in Section \ref{sec:definition}, by discussing existing definitions of CPS in AI along with presenting our definition, and contrasting it to the existing work. We then describe relevant aspects in CC in Section \ref{sec:theory}, and describe how they can be adapted to CPS. In Sections \ref{sec:framework} and \ref{sec:taxonomy}, we present a framework that highlights four key processes involved in CPS. We further classify existing research in the context of the presented framework. In Section \ref{sec:examples}, we discuss examples of CPS architectures in the current literature. Finally, we conclude our survey in Section \ref{sec:future} with open research questions.

\section{Defining Creative Problem Solving in AI}
\label{sec:definition}



We begin by presenting existing definitions of creative problem solving. Derived from the formulation of planning and learning problems in artificial intelligence (which we explain in this section), we introduce our definition for CPS, and highlight its contribution and differences from existing definitions.

\subsection{Existing Definitions of Creative Problem Solving}
\label{subsec:existing_defs}
There have been efforts within the research community to formalize a general definition and process for CPS. Closely related to our work, prior research has formulated CPS based on classical planning and concept re-representation. 




Prior work by Olte{\c{t}}eanu \cite{olteteanu2014two} defines CPS from an object affordance perspective. Affordances broadly refer to action possibilities for objects, e.g., cups are pour-able and doors are open-able \cite{gibson1977concept}. Formally, affordances are defined as relationships between objects, actions, and the effects of applying the actions on the objects \cite{csahin2007afford}. Olte{\c{t}}eanu \cite{olteteanu2014two} distinguishes creative problems from normal problem solving tasks in that CPS problems have poor representational structure, in particular the representation of objects (initial states, relations between objects etc.). Collectively, these entities are referred to as \emph{concepts}, and their representations consist of affordances, visuospatial features, or semantic tags. From this, Olte{\c{t}}eanu defines the process of CPS as \emph{``the ability of a cognitive, natural, or artificial system to use new objects to solve a problem, other than the ones that have been stored in its memory as tools for that specific purpose (if any), or to create those objects by putting together objects or parts of objects the system has access to. Depending on the problem, objects can be either physical or abstract/informational (concepts, problem templates, heuristics or other forms of representations)''} \cite{olteteanu2015seeing}. However, their framework specifically refer to objects, and do not cover other concepts relevant to problem solving in AI, e.g., actions and environment states.


In a contrasting definition, Sarathy and Scheutz \cite{sarathy2018macGyver} define the notion of ``MacGyver-esque'' creativity as embodied agents that can ``\textit{generate, execute, and learn strategies for identifying and solving seemingly unsolvable real-world problems}''. They approach CPS as a planning problem, introducing the notion of a \emph{MacGyver Problem} (MGP) as a \textit{planning problem in the agent’s world that has a goal state that is currently unreachable by the agent} \cite{sarathy2018macGyver}. They formalize MGP with respect to an agent $t$, as a planning problem in the agent's world $\mathbb{W}_t$, that has a goal state $g$ currently unreachable by the agent. In order to solve an MGP, the agent modifies its domain knowledge (through domain expansion or contraction) by sensing and perceiving its environment and its own self, enabling the agent to discover previously unknown information needed for accomplishing the goal. However, the proposed definition of MGPs do not cover learning approaches in AI for CPS and does not describe potential methods for modifying domain knowledge of the agent.



\subsection{Proposed Definition of Creative Problem Solving}

In this section, we present a novel formalization of CPS. We begin by first defining the components of a traditional planning or learning problem to be solved by an agent acting in its environment. 

\subsubsection{Components of Traditional planning and learning in AI} The planning or learning problem specification in AI typically consists of a task goal $G$ to be accomplished, given a set of environment states $\mathcal{S}$ and agent actions $\mathcal{A}$. The agent then produces a solution $\Pi$ for accomplishing the task goal. Depending on whether the initial problem is formulated as planning or learning, the generated solution corresponds to either a task plan or policy respectively, over the states and actions. Thus, $\Pi:\mathcal{S} \rightarrow \mathcal{A}$, represents a mapping from the set of environment states, to the set of actions. In the case of planning, a full task plan is typically computed prior to the agent acting in the environment. Hence, CPS may be observed before any action executes physically. In the case of learning, the policy is typically learned on-the-fly as actions are executed in the environment and CPS may be observed concurrently.

\begin{figure}[t]
	\centering
	\includegraphics[width=0.3\textwidth]{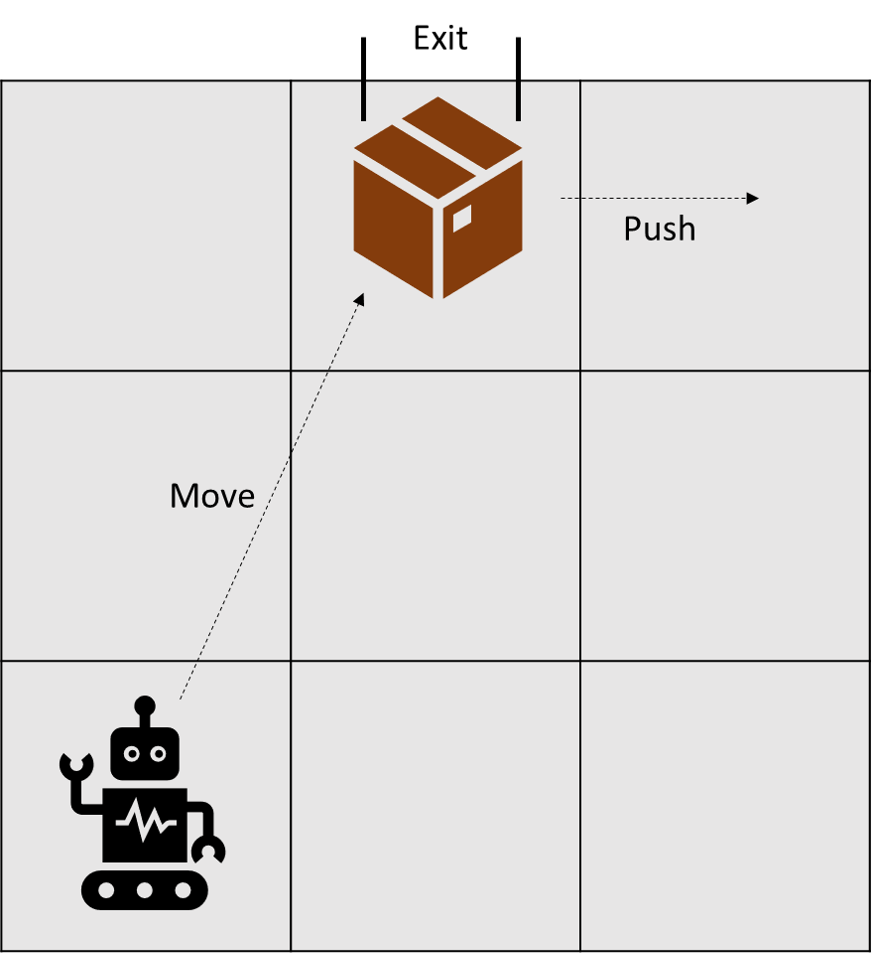}
	\captionsetup{width=\linewidth}
	\caption{Nominal case where creative problem solving is not required, given the knowledge in Table \ref{tab:init-concept-space}. The robot can push the box onto a nearby empty space and proceed to the exit.}
	\label{fig:CPS_definition}
\end{figure}

\begin{table}[t]
\centering
\begin{tabular}{|c|l|} 
\hline
$\bf{C_S}$ & $\bf{c_s^1 \ldots c_s^9}$ = robot at one of $t_{(1,1)}, t_{(1,2)}, t_{(1,3)}, t_{(2,1)}, t_{(2,2)}, t_{(2,3)}, t_{(3,1)}, t_{(3,2)}, t_{(3,3)}$ \\
& $\bf{c_s^{10} \ldots c_s^{18}} =$ box$_{1}$ at one of $t_{(1,1)}, t_{(1,2)}, t_{(1,3)}, t_{(2,1)}, t_{(2,2)}, t_{(2,3)}, t_{(3,1)}, t_{(3,2)}, t_{(3,3)}$ \\
\hline
$\bf{C_A}$ & $\bf{c_a^1}$ = move to a designated location \\ 
& $\bf{c_a^2}$ = push box to a location that is \textit{empty} \\ 
\hline
\end{tabular}
\caption{\label{tab:init-concept-space} Initial concept space of the grid-world agent, shown partially.}
\end{table}

To highlight the definitions presented in this paper, we introduce the following running example in a planning domain. Consider a scenario with a robot in a 2-dimensional 3x3 `grid-world' room with 9 locations denoted by $t_{(1,1)}, t_{(1,2)}, t_{(1,3)}, t_{(2,1)}, t_{(2,2)}, ... t_{(3,3)}$. The room consists of one box in front of an exit door. The robot may have the goal of exiting the room (e.g., G = robot not in room), with two types of available actions, which allow the robot to move around the room, and to move boxes around the room (e.g., A = $\{ a_1, a_2 \}$, where $a_1$ = move to a location, and $a_2$ = push box to an \textit{empty} location). This agent could then generate a solution to get from its current state in the room to the goal state, by using the actions that are known to it (e.g., $\Pi$ = move to the exit, push the box covering the exit door to the empty location on the side, move through exit door). 

\subsubsection{Components of Creative problem solving in AI}
Given the formalisms from planning and learning problems in AI, we now describe the components of CPS. In creative problem solving, we broadly define the notion of a \textit{concept}, as a \textbf{state} (of the environment and/or agent) or \textbf{action}. More specifically, depending on the problem formulation, concepts could refer to the actions that an agent can perform (e.g., ``move''), or the state space including states of objects in the agent's environment (e.g., ``box at location $t_{(1,2)}$''), and the state of the agent (e.g., ``agent at location $t_{(2,3)}$''). Grouping states and actions under the single term ``concepts'' allows us to unify the broad range of problem formulations within a single definition. More generally, we denote concepts as $c_x$, and a \textit{conceptual space} $C_X$ as the set of all concepts $c_x$\footnote{We use $c_x$ only as notational convenience since $c_x$ can refer to either a state $s$ or an action $a$. This allows us to explain our CPS framework across different concepts.}. Here, $X$ (and $x$) relate to either the set of environment states $\mathcal{S}$ or actions $\mathcal{A}$. Hence, $C_S$ denotes the set of all states where each state is denoted by $c_s$, and $C_A$ denotes the set of actions, where each action is denoted by $c_a$. For our grid-world example, the concept space involved is partially highlighted in Table \ref{tab:init-concept-space}. 

Furthermore, let $\breve{C}_{X}$ denote the universal set of the concepts $c_x$, such that $\breve{C}_X$ represents a theoretical conceptual space containing every possible concept that the agent could potentially know about, e.g., $\breve{C}_{A}$ as the set of all possible actions that the agent can perform. In this work, we assume that the initial conceptual space $C_X \subset \breve{C}_X$, i.e., the agent's initial knowledge is limited. Note that $C_X = \breve{C}_X$ is not a practical assumption for real-world agents, since it implies that the agent knows every concept possible. In practice, the agent often encounters problems that it is unable to solve given the initial information available to it. Similar scenarios are commonly observed in robots operating in unstructured environments where they often have to improvise to effectively solve the task. 

\begin{figure}[t]
	\centering
	\includegraphics[width=0.6\textwidth]{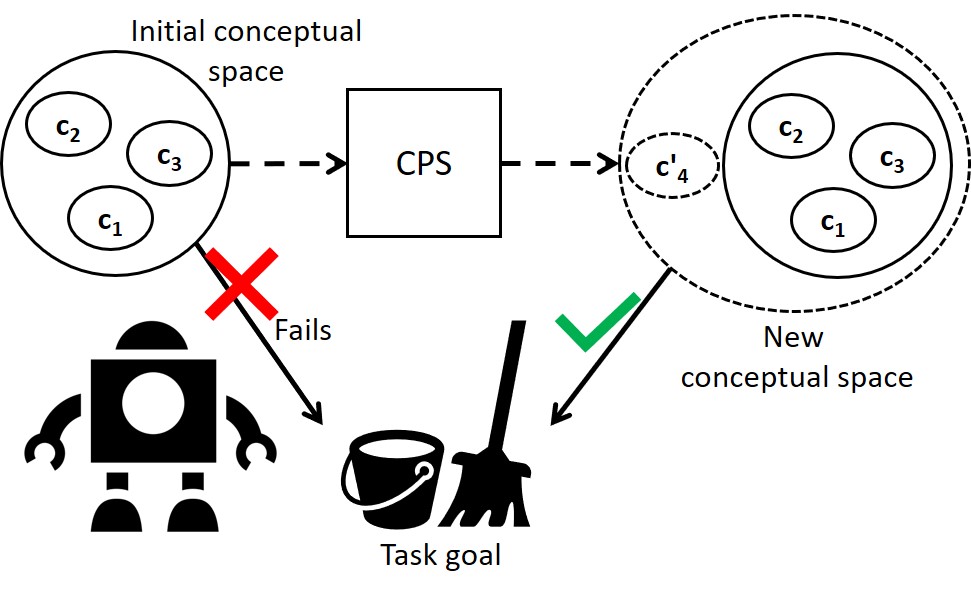}
	\captionsetup{width=\linewidth}
	\caption{Creative problem solving (CPS) occurs when the initial conceptual space of the agent is insufficient to complete the task, and the agent needs to expand its conceptual space to achieve the task goal. Traditional planning/learning approaches in AI would return a failure in such scenarios.}
	\label{fig:CPS_definition}
\end{figure}


\subsubsection{Key definition}
\textbf{A crucial aspect of CPS that differentiates it from general planning or learning problems in AI is that the initial conceptual space $C_X$ known to the agent is \textit{insufficient} to accomplish the task goal}. We refer to such task goals as ``\textit{un-achievable goals}''. Consider the original grid-world example, with an added switch that needs to be \textit{kept pressed} in order to turn off the lights in the room. The new goal for the robot in this case is, G = robot not in room AND lights off. The initial concept space of the robot does not include any action allowing it to accomplish this goal (i.e., Table \ref{tab:init-concept-space} does not contain switch $press$ actions, or switch $is\_pressed$ states). Traditional planning approaches in AI often yield a failure in these circumstances, since the initial set of actions available to the agent is insufficient for completing the task. In traditional learning approaches, the initial action space that the agent is allowed to explore is limited and often not extensible to include new actions. Thus, CPS is characterized by its \emph{flexibility or adaptability} to handle novel problems \cite{guilford1967creativity}. In particular, CPS seeks to enable the agent to discover new concepts for accomplishing the task, by modifying the agent's initial conceptual space. Here, given an initial conceptual space the agent must generate a creative solution $\Pi$ for accomplishing the task. We now present our definition of CPS as follows (also shown in Figure \ref{fig:CPS_definition}):

\begin{definition}
Given an un-achievable goal due to an insufficient conceptual space, creative problem solving is defined as the process by which the agent manipulates its currently known conceptual space in order to discover new concepts that are not in its current conceptual space, thus allowing the agent to accomplish the previously un-achievable goal. Formally, CPS refers to the process by which the agent discovers a new conceptual space $C'_X \nsubseteq C_X$, such that $C'_X = f(C_X)$ is the result of applying some function $f$ on the current conceptual space, enabling the agent to solve the previously unsolvable task by using $C'_X$.
\end{definition}

\begin{figure}[t]
	\centering
	\includegraphics[width=0.6\textwidth]{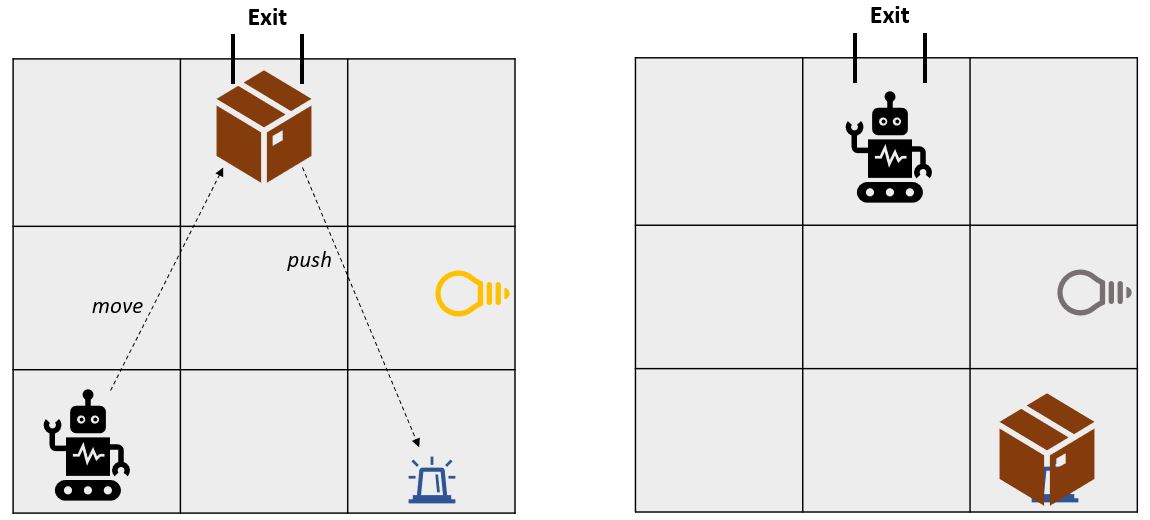}
	\captionsetup{width=\linewidth}
	\caption{Example of CPS in the grid-world planning domain. In the left image, the robot has the goal of exiting the room with the lights turned off. Note that the blue switch needs to be kept pressed in order to keep the lights off. A creative solution in this case involves pushing the brown box over the light switch in order to accomplish the task goal, as shown to the right. The agent here has to somehow discover that the box can be pushed onto the switch to keep it pressed.}
	\label{fig:creative_example}
\end{figure}

In other words, the space of concepts that is explicitly represented by the agent defines the boundaries of what the agent can accomplish. Creativity arises when the agent uses what it already knows to discover something new. In CPS, the newly discovered knowledge is applied to solve a previously impossible task. In our grid-world example with the light switch, the initial conceptual space of the agent (shown in Table \ref{tab:init-concept-space}) is insufficient for accomplishing the goal of <G = robot not in room AND lights off>. However, if the agent is able to somehow discover (via modification of its initial conceptual space) that the box can be pushed onto the light switch to keep the switch pressed, rather than only being able to move it to empty locations, then the agent has exhibited CPS since it can now exit the room while the lights are off (See Figure \ref{fig:creative_example}). More specifically, the agent would have to discover a new action $c^3_a =$ push box onto switch, which has the effect of $is\_pressed($switch$)$. In a learning context, exploration is one possible approach for the agent to discover this new concept, i.e., the agent may push the box around until it happens to be on the switch. This can be computationally prohibitive in larger state spaces. In this paper, we will present three classes of approaches (including \textit{informed} exploration) by which the agent can manipulate the initial conceptual space more efficiently to discover new concepts for accomplishing the task (Section \ref{sec:framework}).



Our definition of CPS differs from previous definitions (Section \ref{subsec:existing_defs}) in the following ways. Firstly, in contrast to the prior definition by Olte{\c{t}}eanu where the concepts focused primarily on object affordances, we describe the notion of concepts in terms of all the core entities involved in a planning or learning problem, i.e., the actions and states (including, but not limited to the states of objects). This allows us to capture CPS approaches that manipulate non-object related concepts (e.g., the agent itself) as well. Secondly, in contrast to the prior definition by Sarathy and Scheutz that formulated macgyvering problems specifically as a planning problem, we present CPS as a planning \textit{or} learning problem, and further connect problem solving in AI to relevant aspects of CC. In particular, we describe i) how CPS can be performed by efficiently manipulating the initial conceptual space via the function $f(.)$ that correlates to existing methods developed in CC, and ii) how the output of CPS can be evaluated by leveraging existing notions of output evaluation in CC. In the following sections, we expand upon our definition, highlighting the theoretical aspects in CC that apply to CPS.

\section{Aspects of Computational Creativity for Creative Problem Solving}
\label{sec:theory}
In this section, we review four major aspects of CC, and their inheritance and adaptations to creative problem solving. The four aspects include: novelty and value, evaluative methods, procedural methods, and Boden's types of creativity. These aspects are grouped into two categories; \emph{output-based aspects} and \emph{process-based aspects}. These categorizations are not meant to divide types of systems, but rather, to group key aspects. We leverage these aspects when presenting the components of CPS in Section \ref{sec:framework}.

\subsection{Output-based Aspects}
\label{subsec:op_aspects}
In output-based aspects, the focus is on evaluating the creativity of a system by determining whether the output produced in a task is considered creative. These systematic outputs, referred to as \emph{artefacts}, may take physical and/or non-physical form (e.g. paintings, songs). The first aspect (\textit{Novelty} and \textit{Value}) describes two key characteristics of a creative output, whereas the second aspect (\textit{Evaluative Methods}) describes methods for evaluating the output.

\boldsec{Novelty and Value} Prior work by Boden \cite{boden1998creativity} proposed that creativity necessitates both novelty and value. Novelty guarantees that the generated outputs of a creative process are original, whereas the value criteria ensures that the generated outputs are not random, but targeted to accomplishing a task goal. Both novelty and value have contextual considerations. An agent may produce a novel painting, but in the context of a scenario which calls for a creative recipe, the novel painting would not be considered valuable \cite{sosa2016multi,varshney2016associative}. 

\boldsec{Evaluative Methods} In evaluative methods, the creativity of a system is evaluated by judging the output of its processes, if it is creative or not. Similar in nature to the Turing test, these methods focus on using the judgement of an external observer. The evaluation can either happen computationally \cite{colton2012computational,varshney2016associative}, from a human evaluator \cite{bishop2010turing,guckelsberger2019addressing}, or from a social group \cite{varshney2016associative}. The nature of these evaluations vary, the output may be compared to a human's creative output, judged in a social context, or evaluated based on the agent's ability to explain its own intentions to a human \cite{cook2019framing}. 

\boldsec{Adaptation of Output-based Aspects in Creative Problem Solving} In creative problem solving, novelty is important with contextual considerations. Creative solutions may not be completely original themselves, but rather in their \emph{application} to the problem. For example, using Tupperware as a container may not be original in itself, but using Tupperware as a replacement for a soap dish may be considered a creative solution to a problem. The second criteria in CC is that creativity necessitates value. In the context of CPS, this criteria is inherited as \emph{usefulness} or \emph{utility}. That is, does the solution actually solve the problem? 

CPS does not directly inherit evaluative methods, because the output of a CPS process is simply evaluated by the agent as either successful or not successful, based on its ability to solve the problem. As such, a successful solution to a problem which necessitates CPS is inherently creative, because problem solving in this case requires the discovery of concepts which are novel relative to the agent. Thus, evaluation in CPS involves evaluating whether the new conceptual space is sufficient to accomplish the current goal.

\subsection{Process-based Aspects}
\label{subsec:process}
Process-based aspects are concerned with the question of \emph{how} creative outputs are produced. The first aspect (\textit{Procedural Methods}) reviews existing methods for synthesizing the creative process, whereas the second aspect (\textit{Boden's Types of Creativity}) reviews three ways of implementing procedural methods. 

\boldsec{Procedural Methods} 
Procedural methods of generating creative outputs consist of two phases: An \emph{expansion} phase where the agent synthesizes a large set of possible outputs for a creative process, and a \emph{contraction} phase where the agent processes the candidate outputs in order to select valuable output. Analogous conceptualizations of the expansion/contraction phases include \emph{divergent thinking/convergent thinking} \cite{guilford1967creativity,zhang2020metacontrol}, \emph{generative thinking/evaluative thinking} \cite{ellamil2012evaluative}, and \emph{defocused attention/focused attention} \cite{sarathy2018real}. 


\boldsec{Boden's Types of Creativity} Boden proposed three ways of generating creative outputs, namely, \emph{combinational creativity}, \emph{transformational creativity}, and \emph{exploratory creativity} \cite{boden1998creativity}. Combinational creativity involves taking known or familiar information, and combining it in a way that generates a novel output. Transformational creativity involves transforming one or more dimensions of the solution/output space to provide the means for new structures to emerge in the transformed space. Lastly, exploratory creativity involves an exhaustive search of a solution/output space to find a novel solution.

\boldsec{Adaptation of Process-based Aspects in Creative Problem Solving} CPS directly utilizes process-based approaches. CPS is triggered by an \emph{impasse} moment, where the agent detects that nominal problem solving techniques are insufficient for accomplishing the goal \cite{knoblich1999constraint}. Impasse is followed by a period of \emph{incubation}, where the agent generates the solution space, synthesizing possible ways of solving the problem using a relaxed representation of the problem and domain. Once a viable solution is found in this space, the agent is said to reach its \emph{insight} or ``Aha!'' moment \cite{colin2019reinforcement}, wherein the agent proceeds to use the solution to solve the problem. We call this process the \emph{impasse-incubation-insight} process. While there exist other general formalizations of the creative process \cite{mumford1991process,mumford1997process}, we use the \emph{impasse-incubation-insight} paradigm for our CPS framework. 

The impasse-incubation-insight process can be implemented using the two part method of expansion and contraction in the following manner -- the impasse moment triggers incubation, where the agent enters the expansion phase and generates a new conceptual space. Upon generating new concepts, the agent enters the contraction phase, wherein the agent applies the newly discovered concepts to generate a plan for accomplishing the goal (insight moment). 
Boden's types of creativity provides three ways to generate the new conceptual space during the expansion phase (referred to as ``Knowledge Manipulation'' in the following sections). We formalize each of these methods in detail in Section \ref{sec:framework}, describing how each method operates on the initial conceptual space.



\section{Components of a Creative Problem Solving Framework}
\label{sec:framework}

\begin{figure}[t]
	\centering
	\includegraphics[width=0.97\textwidth]{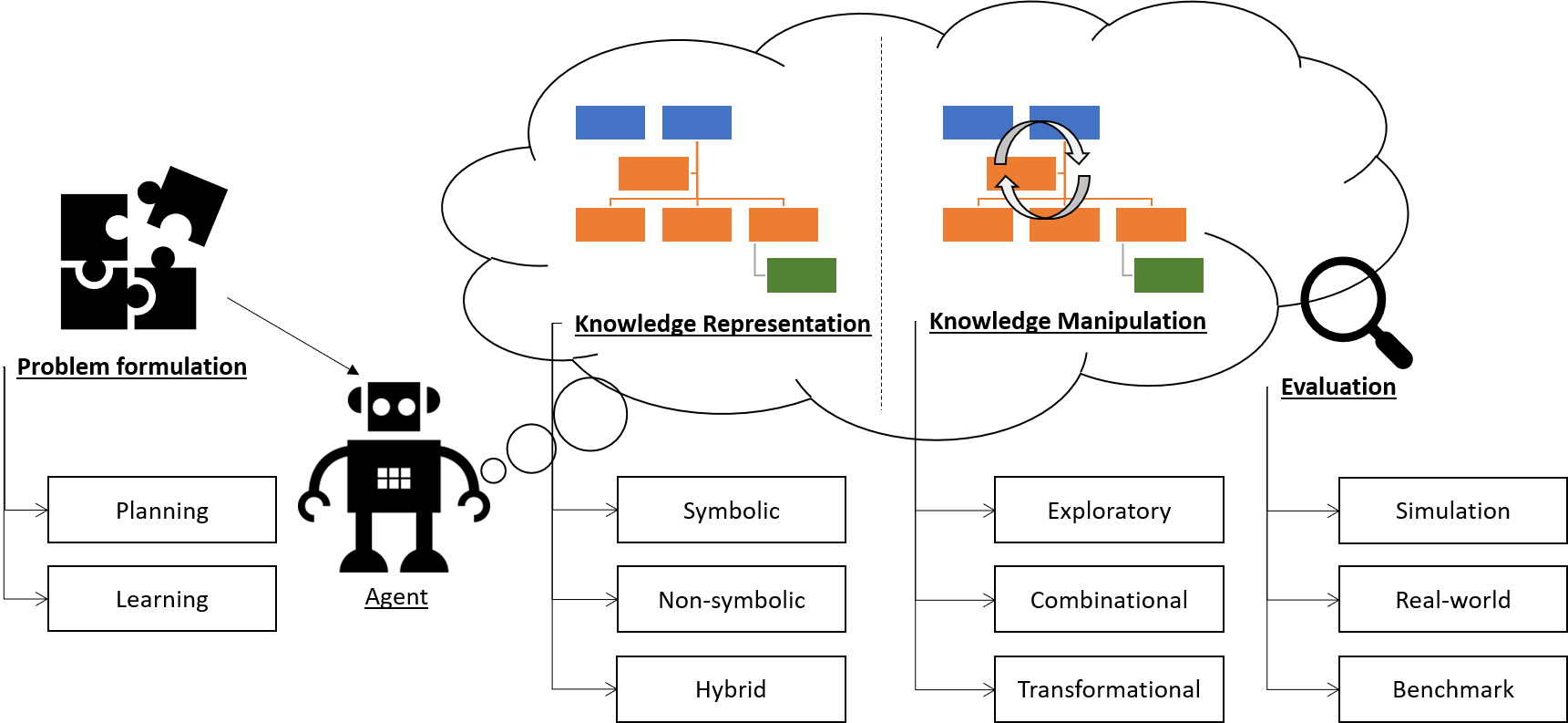}
	\captionsetup{width=\linewidth}
	\caption{Creative problem solving framework, beginning with the problem formulation, followed by representation of the initial conceptual space (knowledge representation). The agent then operates on the initial conceptual space to derive a new conceptual space for solving the task (knowledge manipulation), and evaluates the solutions generated from the new conceptual space for their success (evaluation).}
	\label{fig:abs_taxonomy}
\end{figure}

In this section, we introduce a novel computational framework for creative problem solving (also shown in Figure \ref{fig:abs_taxonomy}), leveraging from the aspects from Section \ref{sec:theory}. Given a task that is currently unsolvable (i.e., impasse), the first step within our framework involves appropriately formulating the problem. In this case, the problem may be formulated as a \textit{planning} and/or \textit{learning} problem, with a few exceptions. Once the problem is formulated, the agent must appropriately represent the relevant information (i.e., the concepts) in order to form the initial conceptual space. In CPS, the initial conceptual space is insufficient for accomplishing the task and as a result, the agent must expand its conceptual space (i.e., incubation) to discover a new conceptual space for accomplishing the goal. The final component of the framework involves evaluating the new conceptual space for its effectiveness in solving the problem, by generating a solution from the new conceptual space (i.e., insight). In summary, we organize existing work in CPS through the following questions: a) \textbf{How is the problem formulated?} (Section \ref{sec:prob_form}: Problem Formulation); b) \textbf{How are the concepts represented?} (Section \ref{sec:know_rep}: Knowledge Representation); c) \textbf{How is the new conceptual space derived?} (Section \ref{sec:know_manip}: Knowledge Manipulation); and d) \textbf{How is the new conceptual space evaluated?} (Section \ref{sec:eval}: Evaluation).

\subsection{Problem Formulation}
\label{sec:prob_form}
\textit{How is the problem formulated?} There are primarily two problem formulations within the CPS literature, namely, a) planning problem, and b) learning problem. In particular, learning refers to reinforcement learning. We categorize and discuss each paper in terms of their predominant methodology. A small subset of the papers in our review do not fall clearly within either problem formulation, which we discuss in detail. 

\subsubsection{Planning} 

\begin{figure}
\centering
\begin{subfigure}{.5\textwidth}
  \centering
  \includegraphics[width=0.9\linewidth]{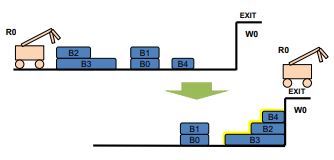}
  \label{fig:sub1}
\end{subfigure}%
\begin{subfigure}{.5\textwidth}
  \centering
  \includegraphics[width=0.5\linewidth]{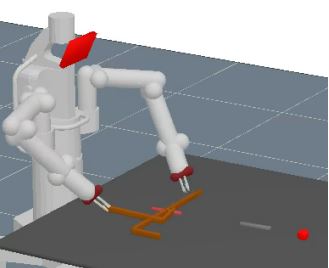}
  \label{fig:sub2}
\end{subfigure}
\caption{Examples of planning in CPS: Task planning within ICARUS \cite{choi2018creating} (left) where the specifics of the motion trajectory is not considered; Task and Motion Planning (TAMP) (right) for sequential manipulation that considers both the task and motion trajectories \cite{toussaint2018differentiable}.}
\label{fig:planning}
\end{figure}

We begin by defining a planning problem in AI. A planning problem consists of a set of states $S$, a set of actions $A$ and state transitions $\gamma$. Further, the formulation consists of an initial state denoted as $s_i$, and a goal state denoted as $s_g$. Most commonly in classical planning, a problem specification consists of a domain definition $\mathcal{P_D} = (S, A, \gamma)$, and a problem definition $\mathcal{P_T} = (\mathcal{P_D}, s_i, s_g)$. Given the domain and problem definitions, planning involves identifying a sequence of actions that can get the agent from the initial state to the goal state (i.e., a task plan). Within CPS, planning formulations may be integrated into a larger architecture or framework, or alternatively presented as standalone approaches. 

\smallsec{Architectural Integration} As an example of architectural integration of task planning, Choi et al. \cite{choi2018creating} introduce planning in the context of a cognitive architecture called \textit{ICARUS}, for the creative construction of navigational structures, e.g., ramps and bridges. In similar work, Freedman et al. \cite{freedman2020} introduce analogical reasoning capabilities into a classical planning architecture called \textit{CIRCA}, to enable a robotic agent to reason about analogies when identifying substitute objects for building navigational structures. In contrast to the construction of navigational structures, prior work by Nair et al. \cite{nair_FGS,nair2021defense} introduce the \textit{Robogyver} architecture that extends classical planning through supervised learning, to enable a robot to create or ``macgyver'' novel tools from available objects. In similar work, Wicaksono and Sammut \cite{wicaksono2017towards,wicaksono2020cognitive} integrate planning within the \textit{CREATIVE} architecture, to enable a robot to craft novel tools through 3D printing, as opposed to using available objects. Lastly, in the cognitive architecture of \textit{SOAR}, Lieto et al. \cite{lieto2019beyond} show that concept representation in a knowledge base can be used as a means for ``subgoal'' resolution (or plan repair) within planning. 

\smallsec{Standalone Task Planning} In addition to planning in the context of cognitive architectures, several standalone planning approaches have also been proposed for CPS. The standalone approaches often focus on either task planning, or motion planning. Task planning involves generating high-level action sequences for accomplishing a task, whereas motion planning focuses on generating sequences of valid joint configurations (of the robot) for performing different actions. In these cases, creativity arises either at the level of the task plan, or at the level of the motion trajectory. 

Within task planning for construction of navigational structures, prior work by Erdogan and Stilman \cite{erdogan2013planning} incorporate constraint optimization as a means of evaluating candidate states during planning. More specifically, they search for specific object configurations (when combining objects to construct structures such as bridges), in a convex continuous domain. For each abstract action, they partition the convex space, and evaluate whether a feasible solution exists in the partitioned spaces. Levihn and Christensen \cite{levihn2015using} extend constraint relaxed planning (using $A^*$) with inverse affordances to enable a robot to navigate its environment in novel ways, e.g., cross gaps by making bridges. Inverse affordances are a mapping from a failed action to object properties that are required to make the action feasible. The agent then locates objects that satisfy the desired properties, and incorporates them into the planner. Prior work by Saboia et al. \cite{saboia2019autonomous} leverage mathematical descriptions of elevation in the terrain to extend planning, and incorporate the construction of ramps to navigate uneven terrains. In closely related work, Tosun et al. \cite{tosun2018perception} use a specialized planner to enable robots to create ramps for navigational tasks. The planner itself serves two functions: 1) it synthesizes a robot controller for achieving the task, and 2) it executes the controller. In contrast to navigational tasks, Boteanu et al. \cite{boteanu2015towards}, focus on using Hierarchical Task Networks (HTNs) for CPS by incorporating novel uses of objects within the planner, e.g., using a bowl instead of a basket. Prior work has also focused on discovery of new actions in the context of planning. Suarez-Hernandez et al. \cite{suarez2020strips} introduce a novel approach for the unsupervised synthesis of new action primitives. Gizzi et al. \cite{gizzi2019creative,gizzi2021toward} discover novel actions through action segmentation and behavior babbling, respectively, which they use as a method for knowledge expansion. In this way, the agent can then re-plan toward a goal in a novel scenario with the new knowledge. In similar work, Sarathy et al. \cite{sarathy2020spotter}, discover action operators through RL as a way to expand the knowledge base for CPS through re-planning. 


\smallsec{Motion Planning} While the approaches described above focus on task planning, several CPS approaches have also been introduced for creative motion planning. Fitzgerald et al. \cite{fitzgerald2017human,fitzgerald2014representing} focus on the transfer of skills (i.e., motion trajectories) from a source object to a target object. For example, their work introduces approaches for adapting motion trajectories for novel uses, such as adapting trajectory for a ladle with a short handle to one that has a much longer handle. In similar work, Gajewski et al. \cite{gajewski2019adapting} focus on adapting trajectories from a source object to a target object by reasoning about the geometric similarities between the source and target objects. Their work differs from that of Fitzgerald et al. in terms of the underlying representations used. In contrast to trajectory adaptation, Qin et al. \cite{qin2020keto} introduce an approach for enabling robots to manipulate novel objects as tools for different manipulation tasks. Here, the robot reasons about tool use based on the underlying object representations, as opposed to adapting a known trajectory from a source to a target object. In contrast to tool manipulation, Murooka et al. \cite{murookaself} focus on incorporating the dynamics of screws and screwdrivers into motion planning, in order to enable a robot to self-tighten loose screws on its body to augment its physical capabilities. \\



Some CPS approaches leverage high-level task constraints for low-level motion planning. For example, Toussaint et al. \cite{toussaint2018differentiable} introduce a novel approach for ``Task and Motion Planning'' (TAMP) to address the problem of sequential manipulation for tool use. They impose explicit task constraints on actions (in a continuous space) regarding the physical dynamics of objects, and optimize over the constraints using motion planning. Similarly, Silver et al. \cite{silver2021learning} show how bottom up relational learning can support learning new probabilistic operators in a TAMP paradigm. Low-level transitions are converted into high level state representations of lifted effects (predicates with argument placeholders), which are then processed via greedy/best-first search to discover new preconditions. Hence, all of the approaches described in this section enable robots to adapt to novel and unforeseen task environments through planning.



\subsubsection{Learning}

\begin{figure}
\centering
\begin{subfigure}{.5\textwidth}
  \centering
  \includegraphics[width=\linewidth]{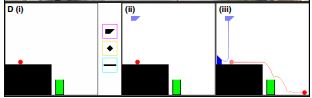}
  \label{fig:sub1}
\end{subfigure}%
\begin{subfigure}{.5\textwidth}
  \centering
  \includegraphics[width=0.7\linewidth]{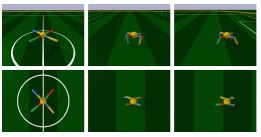}
  \label{fig:sub2}
\end{subfigure}
\caption{Examples of learning in CPS: Agent policy optimization for solving tool-based puzzles \cite{allen2019tools} (left) where the agent design is not considered, and joint optimization of agent policy and design, e.g., morphology of the legs of the agent \cite{ha2019reinforcement} (right).}
\label{fig:learning}
\end{figure}

Within the CPS literature, learning is used in two ways: a) to learn a solution or policy for accomplishing a task goal (i.e., through reinforcement learning), and b) to learn representations that can then be combined with planning or reinforcement learning (RL) techniques. In this section, we focus on the former since it relates specifically to the problem formulation. In section \ref{sec:know_rep}, we focus on the latter, and discuss how learning is applied to learn representations for the conceptual space. 

We begin by presenting the components of an RL problem. An RL problem is typically formalized as a Markov Decision Process (MDP) consisting of the tuple $(S, A, P, \gamma, R)$ that represents a set of states $S$, a set of actions $A$, and transition probabilities $P(s'|s,a) = Pr(s_{t+1} = s' | s_t = s, a_t = a)$, i.e., the probability of transitioning to a state $s'$, given an initial state $s$ and action $a$. The MDP further specifies a discount factor $\gamma$, and a \textit{reward function} $R(s,s',a)$ that indicates a reward for transitioning from $s$ to $s'$ via action $a$. The goal of the agent is to generate a policy (mapping from states to actions) that maximizes the expected reward. In contrast to planning, the agent explores its environment to gather information (via the reward function) to eventually settle on an appropriate policy or solution. In CPS, existing RL approaches have been used to generate creative solutions to tasks. While most of the approaches focus on optimizing the agent's policy alone, a subset of the approaches focus on jointly optimizing for the agent policy \textit{and} the agent's physical design.

\smallsec{Agent Policy Optimization} When optimizing for the agent's policy, Xie et al. \cite{xie2019improvisation} use model-based RL to enable a robot to improvise using tools, e.g., using unconventional tools to grab out-of-reach objects. Similar work by Allen et al. \cite{allen2019tools} introduce the Sample-Simulate-Update-Predict (SSUP) model for solving tool-based puzzles that improve upon more traditional RL techniques. Their work samples actions that operate close to objects (in Cartesian space), and simulates the potential outcomes of the actions using a physics simulator. Based on the simulations and the actual outcomes of executing the actions, the agent eventually generates novel policies for completing tool-based puzzles. Prior work by Baker et al. \cite{baker2019emergent} demonstrated emergent creative tool use by agents playing a game of hide-and-seek. In their work, the agent policies were generated by combining two separate networks: 1) a policy network that learns an action distribution, and 2) a critic network that predicts the discounted feature returns. The policies were optimized by using Proximal Policy Optimization (PPO), and Generalized Advantage Estimation (GAE). In contrast to creative tool use, Bapst et al. \cite{bapst2019structured} generate policies for structure construction via two approaches: 1) Multi-layer Perceptron (MLP) Policy that uses a multi-layer perceptron-based algorithm to output actions or Q-values, and 2) Graph Network (GN) Policy that uses a stack of three Graph Networks to output a policy. 


At the intersection of CPS and psychology, cognitively inspired RL approaches for CPS have also been proposed. Chitnis et al. \cite{chitnis2021glib} use model-based RL to learn a goal/action relational model, where exploration goals are selected based on a novelty measure to optimize the exploration of un-visited state spaces. Here, there is no extrinsic reward function, but rather an intrinsic ``novelty'' motivation. They call this approach a ``goal-literal'' babbling (GLIB) approach. In similar work, Oddi et al. \cite{oddi2020integrating} use the notion of intrinsic motivation for skill learning based on self-generated goals, where the measures driving goal generation are based on competence (composed of novelty, curiosity, exploration, and surprise). Goals are selected based on those which have the highest competence improvement rate. Kralik et al. \cite{kralik2016modeling} apply Q-Tree learning, a variant of Q-learning wherein a hierarchical state representation is learned simultaneously with the action policy. They demonstrate that the Q-Tree model simulates CPS observed in monkeys, thus hypothesizing potential underlying cognitive mechanisms. Similarly, Colin et al. \cite{colin2019reinforcement} develop RL-based algorithms for modeling insight in pigeons, motivated by a real life CPS experiment. They use a basic MDP, along with a CNN to support an actor-critic model for agent behavior, suggesting basic RL as a viable possible mechanism for cognitive CPS. In the context of discovering new actions for accomplishing a task, Kroemer et al. \cite{kroemer2014learning} use RL-based approaches to learn low-level movements corresponding to primitives. In similar work, Jain et al. \cite{jain2020generalization} learn novel actions through unsupervised methods and apply PPO for learning task policies.


\smallsec{Joint Agent Policy and Design Optimization} In contrast to optimizing for agent policies alone, existing work in CPS has also focused on jointly optimizing for agent policy and design. These approaches demonstrate the emergence of novel morphologies that are well-suited to specific environments. Prior work by Pathak et al. \cite{pathak2019learning} focus on self-assembling morphologies wherein individual agents (``limbs'') can combine to form new morphologies. They adopt an RL framework wherein the policy parameters are optimized to jointly maximize the reward for each limb. Prior work by Ha \cite{ha2019reinforcement} uses the REINFORCE algorithm to enable an agent to jointly optimize for its policy and physical design. In closely related work, Schaff et al. \cite{schaff2019jointly} use Proximal Policy Optimization (PPO) as opposed to REINFORCE. In these CPS approaches, the agent learns novel designs that enable effective locomotion in its environment. 


Note that most RL approaches represent information in a strictly non-symbolic manner whereas classical planning often represents information in a symbolic manner. In some cases, the representations used in RL are themselves learned using supervised and semi-supervised learning techniques \cite{xie2019improvisation}. We will describe these in more detail in Section \ref{sec:know_rep}. 

\subsubsection{Other Paradigms}
The papers described in this section conform to our definition of CPS, but do not fall strictly under planning or learning categories, since they do not use the newly discovered concepts to accomplish some task goal. A notable observation is that the majority of the papers in this section pertain to tool-use. They involve the selection of tools that \textit{can be} improvised for a particular function but is not demonstrated in the context of accomplishing some task goal. Hence, we include them in a separate section.

 
Within tool use, prior work has focused on identifying creative alternate uses for objects, through various representations such as graphs \cite{zhu2015understanding,yang2020autonomous}, semantic networks \cite{oltecteanu2016object}, geometric representations \cite{schoeler2015bootstrapping,abelha2016model,shrivatsav2019tool,nair2019autonomous,nair2019tool}, and perceptual functions that capture the effects of interacting with objects using tools \cite{sinapov2007learning,sinapov2008detecting}. In contrast to accomplishing a task goal, the approaches discussed here are evaluated by comparing to ground truth data. For instance, the approach may identify a rock as a good substitute for a hammer, but it is not used to actually perform a hammering task. Rather, it is evaluated against a ground truth label indicating how well rocks can be used for hammering. In this sense, these approaches do not strictly fit a planning or RL paradigm. 

Apart from tool use, the other papers in this section focus on discovering new skills \cite{hangl2017skill,xu2018neural} and discovering new agent designs \cite{zhao2020robogrammar}. Hangl et al. \cite{hangl2017skill} combine already known skills to discover new behaviors which are evaluated for their success at small sub-tasks such as grasping and placement of objects, without a planning or RL formulation. Xu et al. \cite{xu2018neural} leverage Neural Task Programming (NTP) to learn skills by decomposing demonstrated tasks into generalizable substructures. While they demonstrate learning of new skills, they do not conform to the typical planning/RL formulation. In the context of learning new agent designs, Zhao et al. \cite{zhao2020robogrammar} use Model Predictive Control (MPC) to evaluate and optimize for agent designs for traversing different kinds of environments. As a hybrid approach, Sarathy et al. \cite{sarathy2020spotter} combine planning and RL to discover and execute new actions/policies to acquire an environmental states in which the previously unachievable goal can be attained.

\subsection{Knowledge Representation}
\label{sec:know_rep}
\textit{How is the conceptual space represented?} In the previous section, we described the components of a problem formulation (for learning or planning) including states and actions. These specifications form a part of the conceptual space of the agent. In this section, we discuss how the conceptual space (i.e., information regarding actions and states) is represented. In existing CPS works, there are two broad classes of representations that are most commonly used, namely symbolic and non-symbolic. More recent approaches have also sought to combine the two, to develop hybrid approaches that can leverage the relative strengths of both representations. Note that, as described in Section \ref{sec:prob_form}, symbolic representations are typically associated with planning problems, whereas non-symbolic representations are typically associated with learning. In the case of hybrid representations, symbolic and non-symbolic formulations are combined and used for planning and/or learning.

\subsubsection{Symbolic Representations}\label{subsec:sym}
Symbolic representations involve explicitly modeling all known and newly discovered concepts in a declarative form, via facts and rules. These representations encode high-level information as predicates and/or fluents, and the agent often learns relationships between concepts as rules or facts. In this section, we review approaches that use symbolic representations for CPS, including automated planners and large-scale semantic networks.

\smallsec{Planning Languages} Automated planning methods in CPS depend on symbolic representations of states, actions and transitions in the form of planning languages. Two commonly used planning languages include the Stanford Research Institute Problem Solver (STRIPS) language \cite{fikes1971strips}, and the Planning Domain Description Language (PDDL) \cite{mcdermott20001998,mcdermott1998pddl}. Planning languages encode information using logical predicates describing states and actions. States are represented as a list of logical predicates that hold true when the agent is in that particular state. Actions are encoded with \emph{preconditions}, i.e., a list of logical predicates that must hold true before action execution, and \emph{postconditions}, i.e., logical predicates that are expected to hold true after execution. Given an initial state and goal state, planners then use forward or backward chaining of predicates and postconditions to yield a solution. Apart from PDDL and STRIPS, examples of planning languages include PDDL 2.1, which handles temporal planning domains \cite{fox2003pddl2}, PDDL 1.0, which is able to capture domains with probabilistic effects \cite{younes2004ppddl1}, and PDDL 3, a constraints based PDDL planner \cite{gerevini2009deterministic}. Other examples include the Action Description Language (ADL) \cite{pednault1987formulating} and Hierarchical Task Networks (HTNs) \cite{erol1994umcp}.

Erdogan et al. \cite{erdogan2013planning} break down problems in high dimensional continuous spaces into discrete, symbolic PDDL actions used in a constraint-based planner. Choi et al. \cite{choi2018creating} use a STRIPS-inspired planning language extended with numeric representations in order to capture quantitative properties of objects such as their dimensions and weight. Planning and execution is then performed through a modified planner in the ICARUS architecture, which take these quantitative attributes into consideration. Prior work by Wickasono and Sammut \cite{wicaksono2017towards,wicaksono2020cognitive} capture properties of tools within a hierarchical planning language that is then used to create novel tools. Their hierarchy captures aspects such as length, width, and shape of the tools. They capture the state representation at two levels: primitive and abstract. Primitive states contain quantitative values, such as pose of objects, whereas abstract captures qualitative relationships between objects. Suarez-Hernandes \cite{suarez2020strips} introduce a novel algorithm for the synthesis of STRIPS actions from execution traces (sequences of actions), within a planning framework. In Sarathy et al. \cite{sarathy2020spotter}, the agent uses RL in a fully symbolic 2D 'grid-world' domain to resolve failures in PDDL plans. Chitnis et al. \cite{chitnis2021glib} utilize a fully symbolic problem representation, with grounded and un-grounded literal representations of states, actions, and goal states (which they refer to as ``goal-literals''. Using this information, agents ``plan to learn'', where goal-literals drive exploration of novel goal/action combinations to learn a symbolic transition model.

\smallsec{Semantic Networks} In contrast to planning languages, semantic networks are symbolic encodings of information with their inter-connections (e.g. semantic similarity), usually taking the form of a graph structure (also referred to as ``knowledge graphs'' \cite{helie2010incubation}). The connections and their corresponding strengths can be hand-coded, or learned over time. In Lieto et al. \cite{lieto2019beyond}, CPS is performed by exploring conceptual blending in a semantic network of concepts, represented using a non-monotonic logic system which allows for concept composition. This representation accounts a measure for ``common-sense applicability'' on its semantic concepts. In the work of Olte{\c{t}}eanu \cite{olteteanu2014two,oltecteanu2016object}, semantic information (referred to as ``concepts'') is captured based on affordances, visual features, and explicit semantic tags. This collective knowledge base is called a ``semantic map,'' and can assume many different structures. The semantic map is used for object replacement and object composition (OROC) for creating new objects. Boteanu et al. \cite{boteanu2015towards} use large-scale semantic networks, particularly ConceptNet, that capture object affordances in the context of Hierarchical Task Networks (HTNs), to enable a robot to reason about object replacement. In similar work, Freedman et al. \cite{freedman2020} encapsulate knowledge about various object features (such as geometry, length, width, rigidity etc.) within a graphical semantic network to enable a robotic agent to perform analogical reasoning. Prior work by Zhao et al. \cite{zhao2020robogrammar} introduced the notion of a robot ``grammar'' (called ``RoboGrammar'') for generating novel robot designs for environment traversal. They represent the robot's morphology using a graph that symbolically describes the different parts of the robot, such as \textit{head}, \textit{body}, \textit{joints}, \textit{connectors}, and \textit{limbs}. Each node within the graph denotes a particular part, and the edges denote the relationship between the parts.

A key advantage of symbolic approaches (planning languages or semantic networks) is that they provide a vehicle for explanatory reasoning that is human readable and allows the reasoning process to be easily understood and interpreted. However, a major disadvantage of symbolic approaches is that in many cases, knowledge must be encoded a-priori, requiring expertise and domain knowledge which can be difficult for non-expert users. Additionally, although knowledge acquisition in symbolic systems is possible (as in the case of the CPS), it can be challenging to acquire complete and useful information in symbolic systems. 


\subsubsection{Non-symbolic Representations}
\label{subsubsec:non-sym}

\begin{figure}
\centering
\begin{subfigure}{.5\textwidth}
  \centering
  \includegraphics[width=0.8\linewidth]{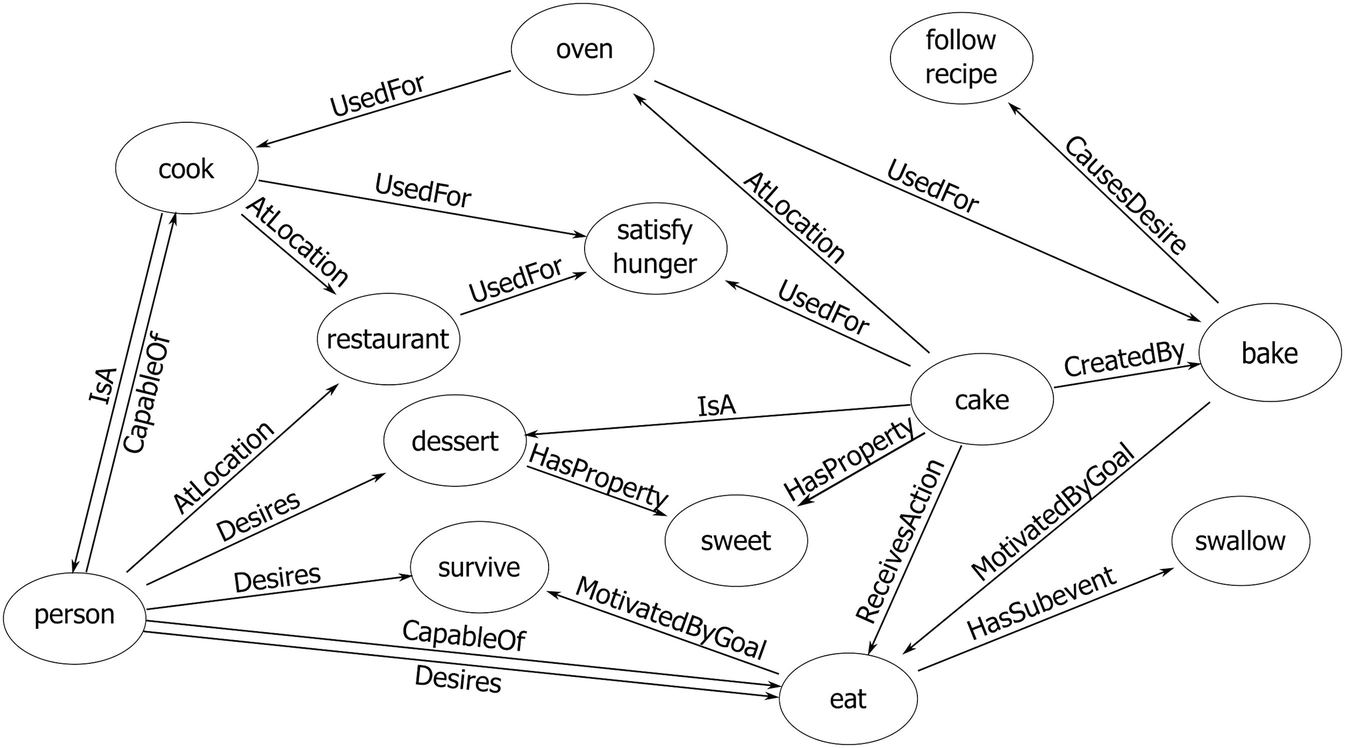}
  \caption{}
  \label{fig:sub1}
\end{subfigure}%
\begin{subfigure}{.5\textwidth}
  \centering
  \includegraphics[width=0.65\linewidth]{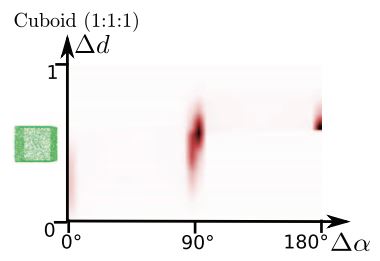}
  \caption{}
  \label{fig:sub2}
\end{subfigure}
\begin{subfigure}{.5\textwidth}
  \centering
  \includegraphics[width=0.65\linewidth]{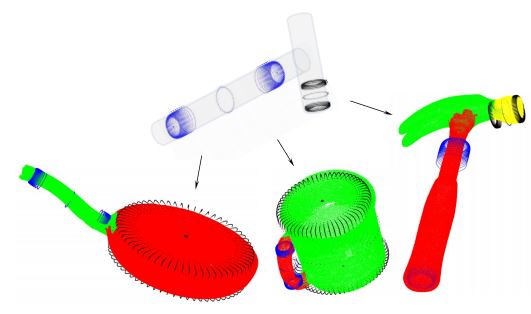}
  \caption{}
  \label{fig:sub2}
\end{subfigure}
\caption{Examples of representations in CPS: a) ConceptNet 5 \cite{speer2013conceptnet}: Symbolically represented semantic networks expressing object-action relationships; b) Non-symbolic non-parametric representation of a cuboid as a histogram \cite{schoeler2015bootstrapping}; and c) Non-symbolic parametric representations of objects using superquadrics (e.g., cylinders representing the handles of containers) \cite{abelha2016model}.}
\label{fig:learning}
\end{figure}

In contrast to symbolic representations, non-symbolic representations do not explicitly model ``rules'' or ``facts''. Instead, non-symbolic representations use \textit{parametric} or \textit{non-parametric} means of representing information. Parametric representations are characterized by the use of numerical parameters that encode some physical meaning. For instance, spheres are characterized by the value of the radius, and the height of a terrain is characterized by the numerical value of its elevation. In CPS, parametric representations are often mathematical models that are pre-specified by the user/developer. In contrast, non-parametric representations cannot be characterized by physically meaningful parameters and are not pre-specified, but rather learned from observed data (called \textit{representation learning}). While pre-defined mathematical models such as models of physics, are used to parametrically represent concepts such as the behavior of objects, non-parametric representations are learned from data via representation learning. In the following paragraphs, we describe these two types of non-symbolic representations.

\smallsec{Mathematical Models (Parametric)}
The mathematical models used in CPS literature are used to generate outputs, and can be thought of as a function $f$ applied to an input $x$. A key distinction between mathematical models and representation learning is that the function $f(x)$ is pre-specified by the user rather than learned from observed data. These mathematical models often capture the underlying dynamics of a system, such as the physics of the environment in order to predict how objects will behave in a given scenario. Within CPS, these models include 3D-transformations, models of the 3D scene or environment, geometric models used to represent objects, mathematical models of environment physics, and inverse kinematic models. 

Prior work by Fitzgerald et al. \cite{fitzgerald2019human} use 3D transformations (rotations and translations) to represent the relationships between tool use trajectories in a source and target environment in order to enable a robot to adapt tool use trajectories learned from human demonstrations in a source environment to a new target environment. In similar work, Abelha et al. \cite{abelha2016model} and Gajewski et al. \cite{gajewski2019adapting} model tools using Superquadrics (SQs) to identify creative substitutes for missing tools. SQs are geometric shapes similar to other quadrics, but raised to arbitrary powers as opposed to the power of two. SQ representations use 13 parameters that characterize various geometric properties of the tools they represent, such as length, width etc. Prior work by Allen et al. \cite{allen2019tools} demonstrate CPS using a Sample-Simulate-Update-Predict (SSUP) model. They model environment physics in order to simulate object behavior in a given environment. The simulations are then used to update a policy for predicting actions that can the enable the agent to solve novel puzzles.  

Within navigation, prior work has focused on deriving parametric representations of the environment by modeling various attributes such as elevation and gaps. Saboia et al. \cite{saboia2019autonomous} model the navigability of an environment by describing mathematical models that capture the height difference between points on the terrain, the pose stability of the robot and, reachability of a particular point on the terrain. These representations are used to enable the robot to construct ramps using available objects in order to traverse the terrain effectively. Closely related work by Tosun et al. \cite{tosun2018perception} introduce a probabilistic template-based terrain characterization algorithm that uses feature templates to identify regions of elevation, given a 2.5D elevation map of the robot's environment. The robot then uses objects that are available to it in order to construct ramps to scale the elevation. Levihn and Stilman \cite{levihn2015using} focus on using objects as simple machines, for enabling a robot to traverse its environment in new ways, such as using a bar of wood to prop open doors. Their work uses an $A^*$ planner defined over a set of constraints in a discretized configuration space of the robot. The constraints model environment physics, such as kinetic energy, momentum, and mass. 

More recent work by Murooka et al. \cite{murookaself} use parametric representations modeled through inverse kinematics to enable a robot to perform augmentation and self-repair by tightening screws on its body. They incorporate the screwdriver and screw dynamics into the inverse kinematic models, including physical concepts such as force, moment, and coefficients of static friction. In some creative results, the robot attaches hooks to its body to enable it to carry grocery bags. Prior work by Hangl et al. \cite{hangl2017skill} parameterize the notions of ``curiosity'' and ``boredom'' using Shannon entropy, in order to guide the learning of new robot behaviors. \\

\smallsec{Learned Representations (Non-parametric)}
Representation learning refers to a set of techniques for learning representations or ``features'' that are useful for encoding a given set of inputs \cite{bengio2013representation}. Here, the term ``useful'' is used to indicate features that are capable of differentiating between classes of inputs. When applied to CPS, non-parametric representations of the initial conceptual space of the agent is learned through observed data. Within the new representation, new concepts are discovered for solving tasks. The vast majority of models used for learning non-parametric representations in CPS include different forms of neural networks, such as Long Short Term Memory networks (LSTMs), Conditional Variational Auto-Encoders (CVAEs), and Feedforward Neural Networks (FNN). In addition to neural networks, few approaches in CPS have also used Support Vector Machines (SVM), k-Nearest Neighbors (kNN), and Decision Trees (DTs). While we do not delve into the details of each machine learning algorithm, we briefly describe the ML approaches used in each paper. \\

\noindent \textbf{\small Supervised Learning:} Among existing CPS approaches that learn non-parametric representations using neural networks, Xie et al. \cite{xie2019improvisation} use LSTMs to learn skill representations for improvisational tool use. In this case, the robot uses data collected from human demonstrations in order to train the LSTMs to represent tool use trajectories for any given tool, often adapting the trajectory to demonstrate improvisational skills. Similar work by Qin et al. \cite{qin2020keto} use the PointNet architecture \cite{qi2017pointnet} for learning keypoint (i.e., points of interest) representations of tools. The network predicts novel keypoints, given an input tool, e.g., Grasping locations for unconventional uses of the tool. Baker et al. \cite{baker2019emergent} use LSTMs within their policy network architecture to enable a set of agents to learn creative policies that involve using objects to succeed in a game of hide-and-seek. Within the space of tool construction, prior work by Nair et al. \cite{nair2019autonomous,nair2019tool,shrivatsav2019tool} have used FNNs and DNNs to perform tool substitution and tool construction. These networks are used to represent objects in terms of their shape and material, in order to identify appropriate objects for constructing or substituting a missing tool. In closely related work, Yang et al. \cite{yang2020autonomous} use Gated Graph Neural Networks (GGNNs) to represent object shapes and reason about pairs of objects for tool construction. Bapst et al. \cite{bapst2019structured} combine two types of internal representations, namely, Recurrent Neural Networks (RNNs) and Convolutional Neural Networks (CNNs) for the construction of physical structures in simulation. Prior work in tool substitution by Schoeler et al. \cite{schoeler2015bootstrapping} uses SVMs to predict alternate uses for given objects. They formulate their work as a part-based affordance learning problem, wherein the models learn shape representations of the tools as histograms (called part signatures) through supervised learning. These representations are then used for identifying tool substitutes. 

Apart from representing tools and structures, neural networks have also been successfully applied to represent the agent. In these cases, the non-parametric representations model the agent's physical design or morphology. Prior work by Pathak et al. \cite{pathak2019learning} have used Dynamic Graph Networks (DGN) in order to enable an agent to reason about its morphology. Here, the limbs and joints of the agent are represented in the form of a graph learned using DGN, and combined with RL to learn motion policies consistent with the graph structure. Similar work, has also used FNNs to represent agent designs, in conjunction with RL approaches such as REINFORCE \cite{ha2019reinforcement} or Proximal Policy Optimization (PPO) \cite{schaff2019jointly}. \\

\noindent \textbf{\small Unsupervised/Semi-Supervised Learning:} Existing approaches in CPS have also used unsupervised/semi-supervised learning methods. Prior work by Sinapov et al. \cite{sinapov2007learning,sinapov2008detecting} represent tools using k-nearest neighbors (kNN) and Decision Trees (DTs). In their work, the robot gathers observations through repeated interactions with various tools and the collected observations are used to cluster and learn affordance representations for new tools. In Kralik et al. \cite{kralik2016modeling}, external state representations are clustered into hierarchical state representations, thus discretizing a continuous non-symbolic state space for RL. Work in generating reusable dynamic movement primitives (DMPs) commonly employ unsupervised learning techniques to extract motion structures for accomplishing tasks. Kroemer et al. \cite{kroemer2015towards} sample non-symbolic state spaces comprised of robot end effector data (end effector torque/position, joint angles, and contact features) and environment data (position, orientation, and relational distances of objects) to learn new representations.

A key advantage of non-parametric representations is that they do not require significant engineering on the part of the user, as they can be learned from data. However, they often lose any meaningful interpretation. In contrast, parametric representations carry some physical interpretation (such as elevation or shape), but require some engineering. In order to leverage the advantages of interpretability and to alleviate hand-coding, several CPS approaches introduce hybrid symbolic and non-symbolic representations.

\subsubsection{Hybrid Symbolic/Non-symbolic Representations}

\begin{figure}
\centering
\begin{subfigure}{.5\textwidth}
  \centering
  \includegraphics[width=0.9\linewidth]{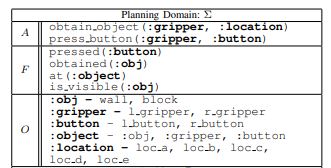}
  \label{fig:sub1}
\end{subfigure}%
\begin{subfigure}{.5\textwidth}
  \centering
  \includegraphics[width=0.9\linewidth]{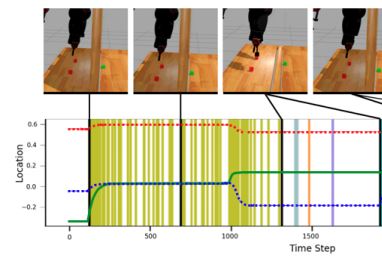}
  \label{fig:sub2}
\end{subfigure}
\caption{Example of hybrid symbolic and non-symbolic representation in CPS: The symbolic PDDL representation of actions (left) is combined with their trajectory representations (action executors) highlighting change points in the trajectory (right) \cite{gizzi2019creative}.}
\label{fig:hybrid}
\end{figure}

Hybrid approaches seek to leverage the strengths of both symbolic and non-symbolic approaches. In most cases, the symbolic representations are used to capture domain knowledge, that is then used to improve learning via non-symbolic approaches. Additionally, the framework may switch between symbolic and non-symbolic representations depending on the environment and the nature of the problem. In Zhu et al. \cite{zhu2015understanding}, videos of tool-use demonstrations are decomposed into non-symbolic representations of objects and trajectories and composed to form a symbolic concept graph, containing spatial, causal, and temporal information of objects. Next, SVM clustering and weight-based ranking algorithms are used to capture the ``essense'' of the task, thereby providing a generative basis for problem representation. As a result, when faced with a new task, the agent is able to use the generative representations to ``imagine'' the use of candidate tools, eventually choosing an appropriate object. The task transfer work presented by Fitzgerald et al. \cite{fitzgerald2017human} uses a previously developed Tiered Task Abstraction (TTA) \cite{fitzgerald2015similarity} framework for abstracting actions using low-level object and motion information, and re-grounding them in novel CPS tasks using both trajectory information and high level object and task descriptors. In Xu et al. \cite{xu2018neural}, actions are called at a symbolic level, where underlying representations exist as low-level object position trajectories relative to the robots gripper frame. In Toussaint et al. \cite{toussaint2018differentiable}, authors propose ``Task and Motion Planning'' (TAMP) that combines symbolic and non-symbolic representations in classical search problems with optimization as a way to accomplish a task. Specifically, they ground a set of symbolic predicates describing dynamic and kinematic contact as physics-based constraint rules, using non-symbolic processing methods. They define a set of modes in terms of these constraints, and perform optimization using a decision tree. Lastly, these constraints are mapped to the action level using a pre-defined knowledge based on symbolic actions, which have constraints encoded in their end effects. In similar work, Silver et al. \cite{silver2021learning} start with using low-level state representations, which are converted into symbolic transitions used for operator learning. This conversion happens through a deterministic parse function, which outputs lossy grounded predicate values (predicates with set arguments). In Gizzi et al. \cite{gizzi2019creative}, non-symbolic methods are used when symbolic planning fails to execute or plan for a creative task, respectively. The CPS agent, upon incurring a plan execution failure, attempts to discover new actions (represented both symbolically and non-symbolically, Figure \ref{fig:hybrid}) through segmenting formerly known actions. In similar follow-up work, Gizzi et al. \cite{gizzi2021toward} utilize a framework for action discovery that applies low-level parameter variations to discover new actions. Low-level parameter variations change symbolic level predicates by either, a) generating a novel effect, or b) generating a set of effects equivalent to the original action, which are then added back into the knowledge base. The representations used here exist in both high-level symbolic form, and a low-level action controller form. Similarly, in Oddi et al. \cite{oddi2020integrating}, a set of 10 continuous low-level state representations are abstracted into predicate form descriptors, to populate the preconditions and effects of newly discovered PDDL operators. This form of skill abstraction is shown to enable problem solving in novel scenarios. Prior work by Nair et al. \cite{nair_FGS} combine symbolic planning with non-symbolic representations learned through supervised learning techniques, to enable a robot to construct novel tools. They use feed-forward neural networks to predict ``scores'' for object-related symbols in the planner, that indicate visual and material fitness of objects for constructing tools. Here, the tool and object shapes are non-symbolically represented, whereas the actions and states are symbolically represented in the planner through PDDL. 

The benefit of implementing hybrid approaches is that they limit the amount of expert knowledge that needs to be provided by the user, while also minimizing the data required to learn non-symbolic representations.

\subsection{Knowledge Manipulation}
\label{sec:know_manip}

\begin{figure}[t]
	\centering
	\includegraphics[width=0.97\textwidth]{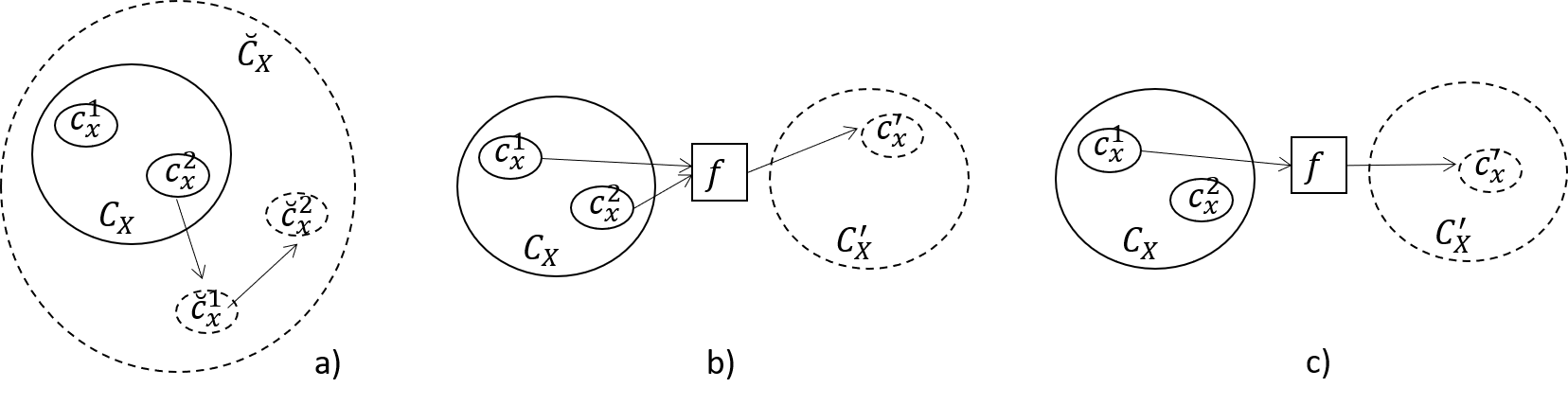}
	\captionsetup{width=\linewidth}
	\caption{Three methods of knowledge manipulation based on Boden's levels \cite{boden1998creativity}: a) Exploratory involving exploration in the universal conceptual space $\breve{C}_X$; b) Combinational where initial concepts are combined to discover new concepts; and c) Transformational where initial concepts are transformed into new concepts.}
	\label{fig:bodens_taxonomy}
\end{figure}

\textit{How is the initial conceptual space manipulated to create a new conceptual space that enables creative problem solving?} In this section, we elaborate on Boden’s three levels of creativity (described in Section \ref{sec:theory}). Specifically, the agent can discover new concepts by: a) exploring the universal conceptual space (Exploratory), b) combining concepts within the initial conceptual space (Combinational), or c) transforming the initial conceptual space (Transformational) \cite{boden1998creativity}. We formalize these approaches (also visualized in Figure \ref{fig:bodens_taxonomy}) in terms of the initial conceptual space $C_X$, and the new conceptual space $C'_X$, and also categorize existing CPS literature accordingly.

\subsubsection{Exploratory Methods}
\label{subsubsec:exp}
In exploratory methods of knowledge manipulation (Figure \ref{fig:bodens_taxonomy}a), the agent searches the universal conceptual space $\breve{C}_X$, in order to discover a new conceptual space $C'_X \subset \breve{C}_X$ and $C'_X \nsubseteq C_X$, that enables the agent to solve the task. Since unguided exploration can be prohibitive, the search is informed by loss/reward functions or heuristics that ``guide'' the agent to explore specific regions of the universal conceptual space. However, this raises an important question, ``\textit{Is all of search or reinforcement learning an example of creative problem solving?}''. In Section \ref{subsec:op_aspects}, we noted the criteria of novelty applied to evaluate the generated output. Re-iterating our ``Tupperware'' example; if the search or exploration yields the solution: ``Use Tupperware as food container'', it is not considered novel, and hence not an example of CPS. Whereas, if it yields the solution: ``Use Tupperware as a soap-dish'', it is considered as an example of CPS. In this survey, we apply this criteria subjectively to identify the subset of the search-RL literature that qualifies as examples of CPS, and we focus our discussion on these papers, highlighting examples of how they satisfy the criteria.


We begin by formalizing exploratory CPS. In the formalization below, the agent uses a loss function, although it can be extended to heuristics and reward functions as well. A newly discovered concept $c'_x \in C'_X$ can be represented as follows:
\begin{gather*}
    \{c'_x = argmin_{\breve{c}_x} \mathcal{L}(\breve{c}_x) \ s.t. \ \breve{c}_x \in \breve{C}_X\},
\end{gather*}



\noindent Here, $\mathcal{L}$ denotes an appropriate loss function, and $C'_X$ contains novel concepts from the universal conceptual space such that $C'_X \nsubseteq C_X$ (Figure \ref{fig:bodens_taxonomy}a). From our grid-world example, the agent may discover how to turn off the lights by pushing boxes around the room guided by an appropriate reward function, in order to eventually discover that it can accomplish the goal by pushing the box on top of the off button. Here, the agent discovers that in addition to moving objects to empty locations, they can be moved on top of some other objects.

\smallsec{Loss/Reward Functions}
Within existing CPS literature, prior work has focused on using cost and reward functions that guide the agent towards creative behavior. Xie et al. \cite{xie2019improvisation} use a cost function that is the expected Euclidean distance between current and goal positions of a target object, in order to guide a robot to use tools in an improvised manner to move the target object from the initial to the goal position. The per-time step costs are summed together and used to select an appropriate tool-use trajectory with the lowest cost. Their work yields novel, i.e., non-prototypical uses of objects previously unknown to the robot, such as using a knife to pull objects closer. Within multi-agent systems, Baker et al. \cite{baker2019emergent} use reward functions for policy learning, wherein the reward functions seek to maximize the total expected discounted return for each agent within the multi-agent population. Their work demonstrates emergent creative tool use, such as using boxes to build a fort to win at hide-and-seek. Prior work by Bapst et al. \cite{bapst2019structured} use reward functions that enable the a set of agents to construct novel physical structures that achieve different tasks such as connecting multiple structures or covering existing structures. 

In the context of discovering creative agent designs, Pathak et al. \cite{pathak2019learning} use reward functions that guide multi-agent systems to assemble into a single agent, leading to the discovery novel agent morphologies. Each self-assembling agent can be considered as a primitive limb with linking actions to join other limbs, and the reward function seeks to maximize the reward for each limb within the joint morphology. The reward itself measures the capability of the joint morphology to perform locomotion tasks. Similar prior work use reward functions within the REINFORCE algorithm \cite{ha2019reinforcement} or Proximal Policy Optimization (PPO) \cite{schaff2019jointly} to enable an agent to optimize for novel agent designs to traverse its environment in new ways. 

\smallsec{Heuristic Search}
Apart from using reward and loss functions, the agent's search may also be guided by heuristics. Heuristic search is commonly observed in the planning literature, wherein the agent uses informed search over a planning space, guided by heuristics, to output a task plan. Prior work by Erdogan and Stilman \cite{erdogan2013planning}, use a pre-defined action ordering (an uninformed heuristic) to decide which child nodes to inspect first during the search. Their work enables a robotic agent to construct novel structures for navigation by effectively searching through the configuration space of the objects. In similar work, Choi et al. \cite{choi2018creating} enable robots to construct makeshift structures (e.g., ramps and bridges) using a unified forward-chaining planner that applies a numeric heuristic to guide a best-first search process. Their heuristic favors states that satisfy more elements in the goal formulae. Levihn and Christensen \cite{levihn2014using}, introduce an approach that samples a set of contact points between objects, and sorts the contact points using a custom scoring function (heuristic). Their work enables robots to use unconventional objects such as planks of wood, or loaded carts to open jammed doors. In Chitnis et al. \cite{chitnis2021glib}, the agent builds a world transition model through state-based goal exploration, where selected goals are optimized on a novelty measure. In the space of tool creation, prior work by Wickasono and Sammut \cite{wicaksono2017towards,wicaksono2020cognitive} use heuristics to guide autonomous tool creation by a robot, where the heuristic prioritizes new tools that are similar to an existing one, where the similarity between tool representations is computed as the number of edit operations needed to transform one representation into the other. Nair et al. \cite{nair_FGS} use supervised learning techniques to compute object fitness scores that are then incorporated into existing planning heuristics within $A^*$ planning to enable a robot to perform tool construction.

In the context of other high-level tasks, prior work that uses heuristics have focused on high-level task planning \cite{boteanu2015towards}, sequential manipulation \cite{toussaint2018differentiable,allen2019tools}, action discovery \cite{suarez2020strips}, and agent design optimization \cite{zhao2020robogrammar}. Prior work by Boteanu et al. \cite{boteanu2015towards} use fitness functions to evaluate candidate objects that can serve as substitutes for a missing object within a hierarchical planning framework (HTN). Their approach to identifying substitute objects proceeds in three steps: generating candidates for the target, extracting contextual information from the HTN, and evaluating the fitness of each candidate within the context. Prior work by Toussaint et al. \cite{toussaint2018differentiable} use the Multi-Bound Tree Search (MBTS) approach to enable a robot to perform creative sequential manipulation tasks using tools. MBTS acts as a best-first search approach, and is applied to a hybrid symbolic and non-symbolic Logic Geometric Program (LGP) to solve for the tool manipulation trajectory. Allen et al. \cite{allen2019tools} use heuristics derived from physics simulators to evaluate agent actions based on their simulated outcomes. The heuristics guide the agent to discover creative policies for solving tool-based puzzles. Suarez-Hernandes et al. \cite{suarez2020strips} introduce cost functions to guide the search for the discovery of new PDDL actions. Their cost function encodes the number of editions required for the new action, such as the number of additions and deletions in effects, as well as changes in the number of preconditions for the actions. Similarly, Gizzi et al. \cite{gizzi2021toward} apply behavior babbling to discover new actions, by varying action controller parameters by partitioning their values evenly along a range of permissible values. For example, given a min and max value of a parameters, of 0 and 100 respectively, a user selected partition value of 5 would cause babbling experimentation for 1, 25, 50, 75 and 100. Oddi et al. \cite{oddi2020integrating} leverage a ``competence''-based intrinsic heuristic for skill learning, where goals are generated to facilitate exploration. Prior work by Zhao et al. \cite{zhao2020robogrammar} use Graph Heuristic Search over a conceptual space of graphs representing robot configurations. The heuristic function is learned as the search progresses, using ground-truth data from MPC-based (Model Predictive Control) evaluations of the robot's performance. 





\subsubsection{Combinational Methods}
In combinational methods of knowledge manipulation (Figure \ref{fig:bodens_taxonomy}b), the agent discovers a new concept $c'_x \in C'_X$ by \textit{combining} existing \textit{distinct} concepts in $C_X$. Hence, the newly discovered concept $c'_x$ can be thought of as a function of distinct concepts $c^i_x$ in the initial conceptual space $C_X$. A key point to note is that, once the new concept $c'_x$ is discovered through combinational manipulation, search or exploration may be used to evaluate or search through the newly discovered conceptual space in order to identify an appropriate solution for completing the task (analogous to the ``Aha'' or insight moment described in Section \ref{subsec:process}). However, the discovery of new concepts itself does not happen through exploration, as is the case with exploratory methods of Section \ref{subsubsec:exp}.


Combinational methods operate as a function over a set of concepts $c_x \in C_X$. Formally, we define a function $f$ that combines $k$ distinct concepts in $C_X$ to discover new concepts\footnote{In cases where more than one distinct concept is not combined, i.e., only a single input is provided to $f$, we define the function to be an identity function, $f(c^i_x) = c^i_x$}, where a single concept $c'_x \in C'_X$ is represented as follows:
\[f:C_X \rightarrow C'_X \ \vert \ c'_x = f(c^1_x, ... c^k_x); c^1_x, ... c^k_x \in C_X, C'_X \nsubseteq C_X\]
Within our grid-world example, combinational methods involve the agent reasoning about combinations of objects, e.g., combination of the box and switch in order to discover that the box can keep the switch pressed. Here, the box-switch combination is a ``composite'' object that is newly discovered, enabling the agent to solve the task. The states of the box-switch composite can be thought of as a combination of the initial concepts regarding the states of the switch and states of the box, wherein the switch ``$is\_pressed$'', if the box is ``$\textit{on}(switch)$''.

\smallsec{Pair-Wise Concept Combination}
Within CPS, some existing works have focused on the use of pair-wise combinations, wherein \textit{two} concepts are combined to discover new concepts. Nair et al. \cite{nair2019autonomous,nair2019tool} introduce the ``Robogyver'' framework for the construction of tools by combining pairs of available objects. They reason about visual properties of objects to output novel tool constructions that combine the properties of the individual objects, e.g., coin + pliers = screwdriver. Here, the visual ``flatness'' of the coin serves as as the head of the screwdriver, while the ``handle'' of the pliers serves as its handle. The novel objects (concepts) are a combination of the visual properties of objects in the initial conceptual space. In closely related work, Yang et al. \cite{yang2020autonomous} reason about object shapes using Gated Graph Neural Networks, that model relationships between pairs of objects for tool construction. The network takes depth maps of the available objects, and a reference tool that needs to be constructed. It outputs the pair of objects that can be combined to best match the provided reference tool. The resultant objects are a combination of the depth maps of the objects in the initial conceptual space. In contrast to reasoning about objects, prior work by Colin et al. \cite{colin2019reinforcement}, show how RL can be used to combine and generalize behavior in agents. In their simulated experiment, they show that after a period of \emph{shaping} (wherein the agent is exposed to a-priori reinforcement-based training), the agent is able to combine two actions ($a_1$ = jumping on a box to peck a banana for a food reward, $a_2$ = pushing a box to a green dot) and generalize their combination to a CPS task ($a'$ = pushing box to location under the banana in order to reach it and peck it for a reward).

\smallsec{Multi-Concept Combination}
In contrast to combining two concepts, existing works have also looked at combining more than two concepts. Prior work by Olte{\c{t}}eanu and Falomir \cite{oltecteanu2016object} focus on composing available objects to create new objects for accomplishing a task (object composition), not limited to pair-wise combinations alone. They introduce the Object Replacement and Object Composition (OROC) framework that combines semantic tags or features associated with objects. Thus the novel concepts are a combination of the semantic features in the initial conceptual space. Closely related is the work of Lieto et al. \cite{lieto2019beyond}, where object substitution is found through combining existing object ``concepts''. Solution objects are selected from a set of candidate objects which are evaluated based on their rank of object similarity to the object being substituted. Prior work by Hangl et al. \cite{hangl2017skill} focus on learning new skills or behaviors as a composition of previously known behaviors. They introduce an approach that enables a robot to discover new behaviors through behavior composition. Given a set of behaviors $B$, new behaviors can be defined as a composition of behaviors $b_i \in B$, as $b_l \circ ... \circ b_2 \circ b_1 \circ b^\sigma$. The goal of their approach is to extend the domain, where the domain is the set of states in which a skill can be applied successfully. The composite behaviors extend domain applicability if they can be applied successfully, i.e., $success(b_l \circ ... \circ b_2 \circ b_1 \circ b^\sigma(e)) = true$. 





\subsubsection{Transformational Methods}
In transformational methods, the agent transforms the initial conceptual space $C_X$ into a new conceptual space $C'_X \nsubseteq C_X$ via some function or ``transform''. The agent then derives the creative solutions from the newly transformed conceptual space $C'_X$. Here, exploratory methods may be used to evaluate or search through the newly discovered conceptual space $c'_x \in C'_X$. However, the discovery of new concepts itself does not occur through exploration.

We formalize transformational manipulation using a transformation function or transform $f$. A concept $c'_x \in C'_X$ can then be represented as follows:
\[f:C_X \rightarrow C'_X \ \vert \ c'_x = f(c_x) \ \forall \ c_x \in C_X, C'_X \nsubseteq C_X\]
Thus, $f$ denotes a surjective function that maps every concept $c_x \in C_X$ to a new concept $c'_x \in C'_X$. Transformational creativity thus involves a mapping from the initial conceptual space to a new conceptual space, via an appropriate transform. In our grid-world example, the agent may transform the initial conceptual space of actions into a new conceptual space that captures the forces applied by the actions. Within this new conceptual space, the agent may discover that pushing a box onto the switch applies a downward force, thus keeping the switch pressed. However, it is a challenging problem to identify the appropriate function for transforming the initial conceptual space into one where the solution for the task becomes evident. This closely relates to to prior work by Olte{\c{t}}eanu et al. \cite{olteteanu2015seeing} (presented in Section \ref{sec:definition}), discussing re-representation within CPS, i.e., thinking about the different representations of concepts wherein particular representations yield the solution to the task. A key point to note that there is a strong correlation between transformational methods and parametric mathematical representations as discussed in Section \ref{subsubsec:non-sym}. Examples of transformations in CPS mostly include geometric and graphical transformations, with some other paradigms. 

\smallsec{Geometric Transformations}
Geometric transformations manipulate geometric aspects of the concepts in the initial conceptual space, e.g., applying rotations and translations. Thus, the concepts are re-represented in terms of their geometrically transformed versions. Prior work by Fitzgerald et al. \cite{fitzgerald2017human,fitzgerald2019human} use 3D rotations and translations in order to re-represent skill trajectories in a new conceptual space where the skills can be re-applied to novel tools. Within the new conceptual space, the robot would be able to adapt previously known skills to new tools that it has not seen before. In similar work, Abelha et al. \cite{abelha2016model}, and Gajewski et al. \cite{gajewski2019adapting} introduce techniques for improvised tool use by mapping the initial conceptual space of tool point clouds to a new conceptual space consisting of their geometric representation using superquadrics. Within the new representation, similarities between tools are utilized to identify how to adapt tool-use trajectories from one tool to another. Sinapov et al. \cite{sinapov2007learning} learn novel tool affordances by transforming the sensory input of the robot (i.e., a 2D image), into five perceptual functions computed over the perceived sensory input. Each of the perceptual functions represent a geometric transform, such as computing image center or gripper location. 

\smallsec{Graph Transformations}
Graph transformations re-represent the initial concepts as a set of features within a graph. Each initial concept is mapped to a feature node in a new graph space. The initial conceptual space may or may not be a graph, but the transformed conceptual space is always a graph. Prior work by Zhu et al. \cite{zhu2015understanding} transform tools into three new graphs: spatial, temporal, and causal parse graphs. Together, they highlight 13 concepts associated with each tool and its use, e.g., material, density etc. Within this new conceptual space, a robot is able to identify appropriate substitute tools for a task. Closely related, prior work by Schoeler et al. \cite{schoeler2015bootstrapping} transform tool point clouds into a set of part ``signatures'' indicating their shape and pose, mapped into a graphical representation of the tool. This graph serves as the new conceptual space for identifying tool substitutes. Prior work by Freedman et al. \cite{freedman2020} represent the initial conceptual space as a graph that captures various properties of objects such as height, weight, rigidity etc. The initial graph is then transformed to a new graphical representation by computing a \textit{maximal common edge subgraph} (MCES) over the initial conceptual space. Within the new graph, the robot performs analogical reasoning to identify appropriate object substitutes for replacing missing objects in novel contexts.

\smallsec{Other Transformations}
Several approaches in CPS have also introduced other transformation functions. Kralik et al. \cite{kralik2016modeling} use complexity reduction to restructure an internal belief representation during the incubation stage of CPS, providing a basis for spontaneous generalization to novelty in problem solving. In navigation, Saboia et al. \cite{saboia2019autonomous} introduce an approach for enabling robots to construct ramps for navigation by transforming the initial conceptual space representing the environment via a custom ``height function'', defined as a mapping from a construction area $Q$ to $\mathbb{R}^+$, as $h : Q \rightarrow \mathbb{R}^+$. In similar work, Tosun et al. \cite{tosun2018perception} map the initial conceptual space representing the environment to a set of ``templates'' that characterize ledges within the environment. The templates are then used by a set of modular robots for the construction of ramps. Similar transformations have also been applied to the action space. Multiple works have shown transformation of actions through abstracting low-level trajectories into higher level behavior models, used for generalization to novel tasks \cite{kroemer2015towards}. Xu et al. \cite{xu2018neural} transform low-level trajectory-based tasks into high level actions by decomposing the tasks to exploit modular substructure. Similarly, prior work by Gizzi et al. \cite{gizzi2019creative} transform the actions (defined at a symbolic level) into a representation of their low-level trajectories, in order to discover new actions via segmentation of the low-level representation. Thus, transforms are applied to the initial conceptual space consisting of high-level symbolic actions, to create a new conceptual space containing their low-level trajectory representation. Prior work has also focused on transforming actions at a strictly symbolic level in order to discover new actions without any change to the underlying low-level mechanics of the actions \cite{sarathy2020spotter,silver2021learning}. Murooka et al. \cite{murookaself} re-represent motion trajectories using inverse kinematic models that account for the physics of screw tightening, such as force, momentum and friction. Within the new conceptual space, the robot was able to repair itself, as well as augment its capabilities by attaching hooks to its body for carrying bags. Lastly, Qin et al. \cite{qin2020keto} re-represent tools using ``Keypoints'' that identify specific points of interest on a given tool point cloud. The tool keypoints include, \textit{grasp point}, \textit{function point}, and \textit{effect point}. The keypoint representation of the tool is then used to derive novel and unconventional tool manipulation trajectories.


With the three modes of knowledge manipulation discussed here, we highlight an important connection between CC and AI, in the context of CPS. Note that existing work in CPS falls into one of the three modes of knowledge manipulation described above, although in the future, CPS frameworks could combine multiple modes of knowledge manipulation. 


\subsection{Evaluation}
\label{sec:eval}

\begin{figure}
\centering
\begin{subfigure}{.5\textwidth}
  \centering
  \includegraphics[width=0.7\linewidth]{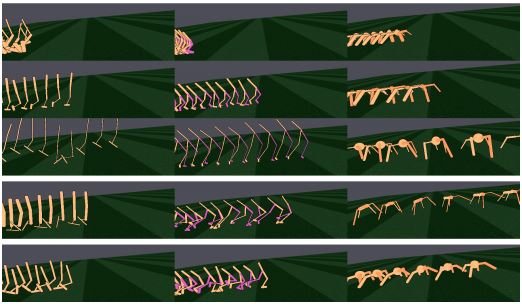}
  \label{fig:sub1}
\end{subfigure}%
\begin{subfigure}{.5\textwidth}
  \centering
  \includegraphics[width=0.7\linewidth]{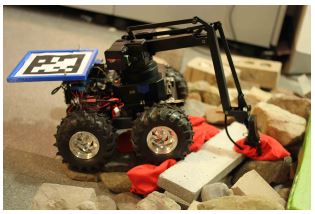}
  \label{fig:sub2}
\end{subfigure}
\caption{Evaluation modes in CPS: Existing physics-based simulators such as OpenAI \cite{schaff2019jointly} (left), and physical evaluation on real robots, including custom platforms \cite{saboia2019autonomous} (right).}
\label{fig:learning}
\end{figure}

\textit{How is the novel conceptual space evaluated?} This section discusses the different modes of evaluation adopted by existing CPS approaches. The mode of evaluation helps distinguish theoretical models from models that have been tested in real-world settings. The classes of evaluation here include, a) using a simulated environment (\textit{Simulation}), b) using a real robot (\textit{Real robot}), and c) developing a standard benchmark test to evaluate the approach.

\subsubsection{Simulation-Based Evaluation}
In this mode of evaluation, the conceptual space is evaluated in a simulated environment. This mode of evaluation is beneficial because it does not require physical access to agents, and can often generate a large number of trials in a small amount of time. Additionally, simulation-based evaluation allows for greater customization capability. Evaluation in simulation happens either as a holistic, end-to-end proof-of-concept evaluation, or as a way to assess a particular feature of a CPS framework.

\smallsec{Physics-Based Simulators}
In proof-of-concept cases, it is typical to use a 2D or 3D physics-based environments, so the end-to-end behavior of the agent can be visually and algorithmically validated. Common examples of such environments include ``Gazebo'' \cite{gizzi2019creative,levihn2015using,toussaint2018differentiable,gizzi2021toward,oddi2020integrating}, ``Unity'' \cite{bapst2019structured,pathak2019learning}, ``OpenAIGym'' \cite{wang2019paired,ha2019reinforcement,schaff2019jointly}, ``Mujoco'' \cite{baker2019emergent}, ``DeepMind Lab'' \cite{leibo2019autocurricula}, ``PyBullet'' \cite{qin2020keto,chitnis2021glib,silver2021learning}, and ``Bullet Physics Library'' \cite{zhao2020robogrammar}. While all of these environments provide similar support, each cater well to different CPS cases. Gazebo is known for its realistic 3D graphics whereas Unity is known for its speed, especially for data-heavy CPS applications that use RL. OpenAIGym is a platform that is typically known for its broad range of RL problem domains, provided off-the-shelf. These domains range from text-based problems, to simple 2D problems, to 3D robotics problems (however these more complex domains are somewhat limited compared to Gazebo and Unity). The open-sourced DeepMind Lab platform provides support for game-based environments as a method for CPS evaluation. Both OpenAIGym and DeepMind Lab provide game-based evaluation environments. Other examples of less common simulation environments include ``DART'' \cite{levihn2014using}, ``CREATE'' \cite{jain2020generalization}, and ``BREVE'' \cite{sinapov2007learning}.


\smallsec{Custom Simulators}
In the case of validating specific aspects of a CPS approach, it may be more favorable to use a specialized or customized simulation environment for evaluation. In simple cases, validation may happen using basic RGB imaging in a 2-D world \cite{colin2019reinforcement,sarathy2020spotter}. In more complex cases, computer vision algorithms built into the simluators are used to identify entities in a physical environment, which are then intelligently processed by a CPS method \cite{zhu2015understanding,abelha2016model,schoeler2015bootstrapping}. Here, CPS algorithms reason about the world without actually interacting with the physical environment. In a more elaborate case, Wang et al. \cite{wang2019paired} focus on automatic or AI-generated testing environments, using the resultant performance to automatically improve the system with a feedback loop. Additional approaches in CPS have also evaluated their methods in simulation, not using a specific simulator, but rather evaluating their approach in terms of simulated metrics \cite{suarez2020strips,freedman2020}.

While simulation-based evaluation is less precise than real-world dynamic environments, many approaches in CPS have been evaluated in this manner, owing to the ease of generating and testing in a variety of environments.

\subsubsection{Real World Evaluation}
In this mode of evaluation, the approaches are evaluated on a real, physical robot. This mode of evaluation can be advantageous in that they can evaluate performance of the approaches against real-world dynamics and noise models.

\smallsec{Robots}
Several existing works in CPS have focused on evaluating the models on various physical platforms such as Sawyer \cite{xie2019improvisation,xu2018neural}, Baxter \cite{wicaksono2017towards,wicaksono2020cognitive,yang2020autonomous}, Kinova \cite{fitzgerald2017human,fitzgerald2019human,nair2019autonomous,nair2019tool,nair_FGS}, PR2 \cite{gajewski2019adapting,murookaself}, KUKA \cite{hangl2017skill}, Darias \cite{kroemer2015towards}, and custom platforms that are specifically built for the tasks that the robot is required to perform \cite{saboia2019autonomous,boteanu2015towards,tosun2018perception,hangl2017skill}. The Sawyer and Kinova platforms each have a single 7-DOF robot arm with a stationary base. Hence, they are not well suited for mobile manipulation applications. The PR2 and ATLAS platforms each consist of two 7-DOF robot arms for bi-manual manipulation, while also supporting navigation. Both the Kinova and Sawyer platforms are quite commonly used for approaches that involve learning from demonstration (LfD) since the robot arms are easily manipulated by humans, particularly when mobility is not a requirement. The custom platforms used in \cite{saboia2019autonomous,boteanu2015towards} consist of robot arms that are attached to a mobile platform in order to enable mobile manipulation, in addition to specific design aspects that make the platform well suited for the task at hand, such as navigation. For example, the custom platform used in \cite{saboia2019autonomous} is built to handle highly uneven terrains. The custom platform used in \cite{hangl2017skill} consists of two KUKA robot arms that are attached to Schunk SDH grippers in order to perform bi-manual manipulation of objects. The custom platforms developed in \cite{tosun2018perception} consist of populations of small modular robots that can attach to one another, with navigation capabilities. 

\smallsec{Sensors}
In terms of sensing capabilities of the platforms, the vast majority of evaluations predominantly test the approaches in unimodal settings (i.e., using a single sensor). These approaches often use only RGB-D sensors that capture partial and noisy point clouds. A small subset of CPS papers evaluate the approaches in multi-modal settings as well, such as using spectrometers in addition to visual inputs \cite{nair2019autonomous,nair2019tool}. Multi-modal sensing capabilities can be especially useful for CPS, since the robot may have to reason about multiple modalities of the objects, including weight, materials, forces etc., rather than solely relying on visual properties, in order to derive creative solutions.


Despite the benefits of evaluating CPS approaches in real-world settings, evaluation on physical robots tend to be slower than simulators, due to hardware related shortcomings.

\subsubsection{Benchmark} \label{subsec:benchmark}
In this section, we discuss papers that introduce benchmarks for evaluating CPS algorithms. Benchmarks allow a standardized comparison of different methods. Designing systems to evaluate creativity often depends on the subjective definitions of creativity proposed by the designer, and more general benchmarks of CPS are yet to be developed. A universal benchmark of CPS would allow a uniform and fair comparison of different approaches. We discuss this limitation in greater detail in Section \ref{sec:future}. Note that a subset of the papers in this section only introduce benchmarks, but not a specific CPS approach itself. However, we include these papers here for completeness, discussing them in this section only.

Existing benchmarks in CPS typically evaluate specific types of creative tasks, rather than a test of general creative intelligence, e.g., benchmarks that focus specifically on creative tool use \cite{allen2019tools}. In many cases, the CPS system is evaluated using an output-based approach which compares the output to that of a human participant. In the case of the ``Alternative Uses Test,'' a CPS system is evaluated in its ability to generate alternative uses of an object by bench-marking its output against the set of corresponding alternative uses generated by a human participant \cite{guilford1967nature}. This benchmark has been used specifically in evaluating CPS systems which employ combinational creativity \cite{oltecteanu2016object,lieto2019beyond}. In some cases, the percentage of CPS problems successfully solved in a given set is used in bench-marking, either in comparison to human performance, or in a standalone manner. Bisk et al. \cite{bisk2019piqa}, propose a benchmark to evaluate common-sense physical reasoning capabilities for CPS by comparing the percentage of accurate reasoning cases to human performance. Kralik et al. \cite{kralik2016modeling}, compare the performance of their Hierarchical RL system to empirical data generated from trials of CPS in Rhesus monkeys. Guzdial et al. \cite{guzdial2018creative}, propose a cross domain metric for evaluating the ability of a CPS system to obtain a goal state, where given a specific domain, an ``Uncreative Max'' (UM) baseline is developed to represent a solution most similar to the desired goal state, given CPS was not employed. Creativity is then measured as any positive score differential between a creative agent and its corresponding UM score. The challenge with this method is in the domain specific crafting of the UM baseline, which is shown to have variance across different domains. This metric is also limited in that it is specific to combinational creativity. 

Benchmarks that extend existing simulation software have also been proposed. The ``POET'' (Paired Open-Ended Trailblazer) benchmark \cite{wang2019paired}, modifies the ``Bipedal walker hardcore'' in OpenAI Gym to create challenging traversal environments for a bipedal walker, requiring the generation of creative traversal policies. Similarly, ``NovelGridWorlds'' \cite{goelnovelgridworlds} present an OpenAI Gym environment framework for evaluating agents that can adapt to sudden novelties in their environments. NovelGridWorlds includes crafting tasks for a bow and pogostick, that requires collection of different resources and presents the AI agents with different classes of novelties, such as object, attributes and action novelties.

Current CPS benchmarks exists solely in simulation, limiting the practical applicability of the methods. Designing benchmarks that evaluate different domains of creative problem-solving, and can be tested on real-robots, remains an open problem (see Section \ref{sec:future}).

\section{Types of Conceptual Spaces}
\label{sec:taxonomy}

\begin{figure}[t]
	\centering
	\includegraphics[width=0.98\textwidth]{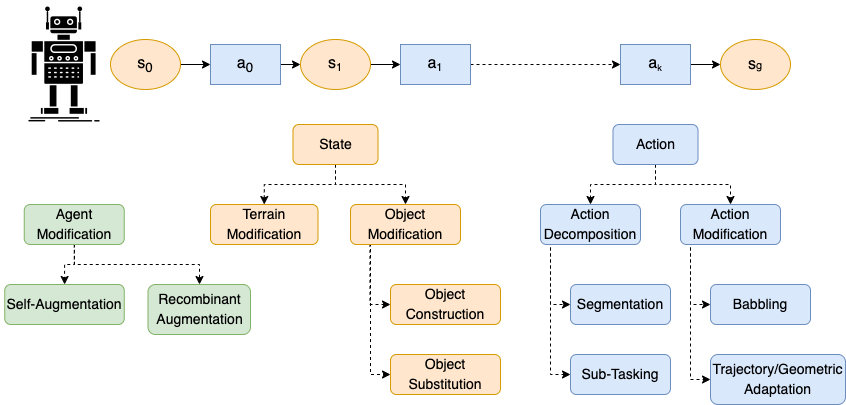}
	\captionsetup{width=\linewidth}
	\caption{The conceptual spaces of the agent that are manipulated for CPS can relate to states or actions. The state space can involve concepts regarding the environment (terrain modification and object modification), or concepts regarding the agent (agent modification). Similarly, manipulating the action space can involve babbling in the existing action space, or modifying existing actions.}
	\label{fig:taxonomy}
\end{figure}

In the previous sections, we discussed the overall CPS framework and presented three approaches for discovering a new conceptual space. However, \textit{what specific information does the conceptual space contain?} As described previously, the conceptual space can be associated with states or actions. While all problem formulations consist of states and actions, either the action space or the state space is manipulated to discover new actions or states. In this section, we categorize existing CPS literature on the basis of the information that is manipulated (also shown in Figure \ref{fig:taxonomy}). 


\subsection{States}

Based on our review, we categorize the existing literature in CPS based on the types of states that are manipulated. These include, a) states of objects in the environment (object modification); b) states of the environment terrain, e.g., holes or ledges in the environment (terrain modification); and c) states of the agent itself (agent modification). In these cases, creativity arises from modifying concepts related to objects, concepts related to the terrain (i.e., modification of the terrain), and concepts related to the agent's design or morphology.


\subsubsection{Object Modification} The CPS literature presented in this subsection involves \textit{modeling and reasoning about the objects themselves}, such as their visual and material properties. Hence, the conceptual space manipulated in these approaches is the states of objects in the environment. Commonly, these problems are referred to as ``Macgyvering'', defined as ``solving problems creatively using whatever objects are available at hand'' \cite{dictionary1989oxford}. Macgyvering as a sub-class of CPS has been researched in both AI \cite{oltecteanu2016object,sarathy2018macGyver} and Robotics \cite{erdogan2016autonomous,nair2019autonomous,nair2019tool}. Existing CPS research in object modification can include a) object substitution that involves adapting objects for non-prototypical uses, e.g., using a pan as a hammer, or b) object construction that involves creating new objects, e.g., constructing a hammer from a rock and stick.

\smallsec{Object Substitution}
Existing research in object substitution often involves comparing the physical attributes of the available objects to those of the missing object in order to identify potential substitutes. Prior work has focused on modeling objects in terms of geometric shapes \cite{abelha2016model,gajewski2019adapting} or shape histograms \cite{schoeler2015bootstrapping}, and reasoning about the objects by computing their similarities to the missing object that is being substituted. The similarity is computed based on the difference between the geometric parameters or a score computed from the shape histograms. In contrast to using geometric similarities alone, Shrivatsav et al. \cite{shrivatsav2019tool} compare both material and shape properties of objects to identify good substitutes. Prior work has also focused on modeling and reasoning about objects via semantic networks representing object properties such as affordances and visual properties \cite{boteanu2015towards,oltecteanu2016object,lieto2019beyond,freedman2020}. Zhu et al. \cite{zhu2015understanding} present a ``hybrid'' approach that reasons about visual properties of objects as well as their physical attributes like mass and density to identify substitute tools, and to suggest appropriate tool-use trajectories for the substitutes. In similar work, Levihn et al. \cite{levihn2015using,levihn2014using} reason about weight and mass in order to identify objects that can supply the required amount of force, or support the desired amount of weight in order to accomplish navigational tasks.

\begin{figure}
\centering
\begin{subfigure}{.5\textwidth}
  \centering
  \includegraphics[width=0.8\linewidth]{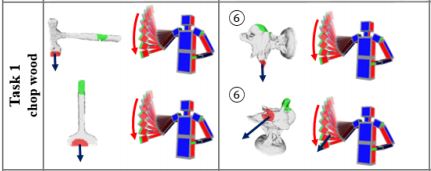}
  \label{fig:sub1}
\end{subfigure}%
\begin{subfigure}{.5\textwidth}
  \centering
  \includegraphics[width=0.7\linewidth]{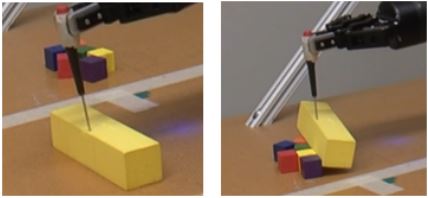}
  \label{fig:sub2}
\end{subfigure}
\caption{Examples of object modification in CPS: Substitution where robot uses non-prototypical objects to chop wood \cite{zhu2015understanding} (left), and construction where robot constructs tools by combining individual objects, such as a screwdriver and foam block, to make a squeegee \cite{nair2019autonomous} (right). In both cases, the agents reason about object properties, e.g., shape and materials.}
\label{fig:learning}
\end{figure}

\smallsec{Object Construction}
In contrast to object substitution, object construction involves combining available objects to construct new objects with desired capabilities. These approaches often involve reasoning about individual objects that can be joined to create a new object that has a combination of their individual properties. There is currently very limited research in the area of object construction. The OROC framework \cite{oltecteanu2016object} proposed by Olte{\c{t}}eanu and Falomir reason about ``object composition'' using semantic knowledge. In this case, the semantic concepts encapsulate affordances of objects, enabling the agent to reason about individual capabilities of objects, e.g., matchbox as a container, and tacks as an attachment method. These can be combined to create a composite object with a new capability, e.g., combining the matchbox and tacks to create a container that can be attached to the wall. In contrast to reasoning about semantic attributes, prior work has also focused on combining objects through visual reasoning alone \cite{yang2020autonomous} or visual and material reasoning \cite{nair2019autonomous,nair2019tool,nair_FGS}. These approaches decompose a reference tool into sub-parts, reasoning about objects that are similar to each sub-part and can be combined to create the reference tool. Similar work by Erdogan and Stilman \cite{erdogan2016autonomous} focus on the autonomous construction of simple machines, further incorporating reasoning about physical concepts such as mass and weight. Wicaksono and Sammut \cite{wicaksono2017towards,wicaksono2020cognitive} focus on creating novel tools from polymers through 3D printing by encoding visual attributes of the tool such as length of the handle or angle of the hook (at the end of the tool). In all of the cases described above, the manipulated conceptual space specifically encodes properties of the objects.

\subsubsection{Terrain Modification} 

\begin{figure}
\centering
\begin{subfigure}{.5\textwidth}
  \centering
  \includegraphics[width=0.6\linewidth]{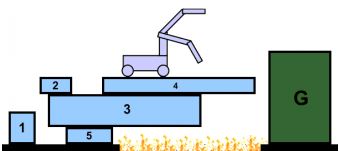}
  \label{fig:sub1}
\end{subfigure}%
\begin{subfigure}{.5\textwidth}
  \centering
  \includegraphics[width=0.5\linewidth]{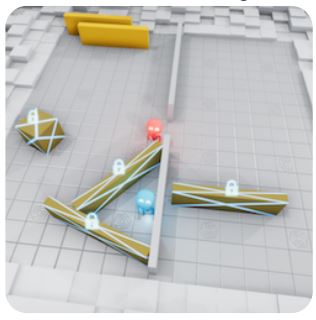}
  \label{fig:sub2}
\end{subfigure}
\caption{Examples of terrain modification in CPS: Example of ADFS where the agent constructs a stable structure to traverse a gap in the terrain \cite{erdogan2013planning} (left), and terrain modification in hide-and-seek where the blue agents build a fort to hide from red agents \cite{baker2019emergent} (right). In contrast to object modification, both these cases involve reasoning about environment properties, such as gaps and walls, as opposed to reasoning about the objects themselves.}
\label{fig:learning}
\end{figure}

The CPS literature discussed in this section involves \textit{modeling and reasoning about the terrain}, e.g., modeling and reasoning about gaps or elevations in the terrain that the agent is unable to cross. Hence, the manipulated conceptual space involved in these approaches is the state of the terrain. Terrain modification often involves modifying unstructured environments to facilitate navigation. Note that terrain modification \textit{can} involve using objects in the environment to modify the terrain. However, the key distinction between terrain and object modification is that the newly discovered concepts relate to the state of the terrain rather than some object in the environment. Further, terrain modification includes approaches that explicitly reason about terrain attributes such as elevation.


\smallsec{Navigational Tasks}
In the context of terrain modification for navigational tasks, ``Automated Design of Functional Structures'' (ADFS) \cite{erdogan2013planning} deals with construction of structures such as bridges to improve terrain navigability. Prior work in ADFS has focused on visual reasoning about the height of elevation points on the terrain \cite{erdogan2013planning,tosun2018perception} and symbolic reasoning of terrain properties such as the widths of a gap \cite{choi2018creating} for traversing the environment in novel ways. More recently, Saboia et al. \cite{saboia2019autonomous} focus on the construction of makeshift ramps using compliant bags, by modeling and reasoning about the reachability or navigability between any two given points on the terrain, in addition to other factors such as terrain elevation.


\smallsec{Non-Navigational Tasks}
Beyond ADFS, terrain modification has also been applied to other domains such as games \cite{baker2019emergent}, and construction of structures for purposes that are non-navigational \cite{bapst2019structured,colin2019reinforcement}. Prior work by Baker et al. \cite{baker2019emergent} demonstrated terrain modification using available objects within a multi-agent system, where the agents attempt to succeed in a game of hide-and-seek. These agents reason about the environment, e.g., walls, to devise strategies such as fort building to \textit{create} unreachable locations in the terrain (as hiders), or to \textit{traverse} unreachable locations to find other agents (as seekers). Prior work by Bapst et al. \cite{bapst2019structured} construct physical structures that reason about, and modify the terrain, for achieving different goals such as connecting separate structures or covering existing structures. In the CPS task of Colin et al. \cite{colin2019reinforcement}, the agent learns to move a box to a certain location in order to obtain a previously unreachable reward. Lastly, in Kralik et al. \cite{kralik2016modeling}, the agent restructures its representational belief about its environment as a means for CPS. Note that in the mentioned non-navigational cases, agents \textit{implicitly} reason about the terrain in the context of reward functions, in contrast to the approaches in ADFS that explicitly model the terrain attributes.

\subsubsection{Agent Modification} 

\begin{figure}
\centering
\begin{subfigure}{.5\textwidth}
  \centering
  \includegraphics[width=0.6\linewidth]{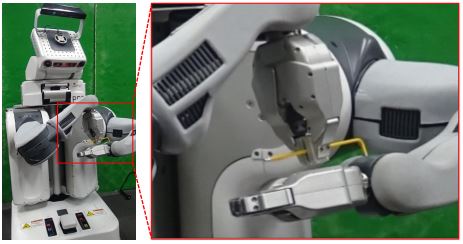}
  \label{fig:sub1}
\end{subfigure}%
\begin{subfigure}{.5\textwidth}
  \centering
  \includegraphics[width=\linewidth]{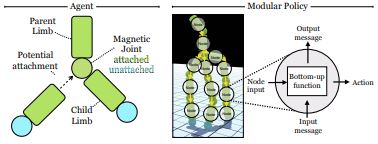}
  \label{fig:sub2}
\end{subfigure}
\caption{Examples of agent modification in CPS: Self-augmentation where a single agent augments its own capabilities, e.g., through self-repair and self-extension \cite{murookaself} (left), and recombinant augmentation where individual agents (``limbs'') combine to form a composite agent, in which case both policy and agent design are optimized \cite{pathak2019learning} (right).}
\label{fig:learning}
\end{figure}

The CPS literature presented in this subsection involves approaches wherein the \textit{agent models and reasons about itself}, e.g., modeling and reasoning about the joints of the agent's body. Hence, the conceptual space involved in these approaches is the state of the agents' body. Depending on whether the approach reasons about a single agent or multiple agents, we divide the existing literature into two classes: Self-augmentation (for single agents), and recombinant augmentation (for multiple agents). Note that in most cases, creativity in agent modification arises from discovering novel agent designs. 

\smallsec{Self-augmentation}
Self-augmentation involves cases where a single agent augments or modifies its own body, in order to accomplish the task. Murooka et al. \cite{murookaself} describe a novel algorithm for self-repair and self-extension of robots. In this work, they model the physical body of the robot using CAD to enable the robot to reason about itself, further utilizing inverse kinematic models of the joints of the robot. By applying these models, the robot is able to tighten screws located on its own body, either for self-repair or to augment its capabilities by attaching hooks to enable them to carry more bags. Prior work by Ha \cite{ha2019reinforcement} jointly optimize the agent's design and policy for navigation tasks. They model various properties of the agent's morphology such as mass, and the orientation of the agent's body parts and joints. Each of these properties are parameterized and incorporated into the policy network. In similar work, Schaff et al. \cite{schaff2019jointly} parameterize the lengths and radii of the links within the robot's configuration, wrapped into an optimization function that is used to guide the policy learning. In contrast to modeling agents within an RL framework, prior work by Zhao et al. \cite{zhao2020robogrammar} introduce ``RoboGrammar'', that models robot designs as a graph. These graphs represent various joints, body parts, and connectors, that make up the configuration space of the robot. 

\smallsec{Recombinant Augmentation}
Recombinant augmentation involves populations of agents that combine in order to augment the capabilities of the individual agents by reasoning about their individual as well as collective capabilities. Existing research has looked at modular agents that combine to create a new \textit{single agent} with capabilities that the modular agents do not possess individually. Specifically, prior work by Pathak et al. \cite{pathak2019learning} have looked at self-assembling morphologies, where individual agents combine to create a composite agent that is capable of efficiently navigating its environment. Each primitive agent in the population can be construed as a ``limb'' wherein the limbs may choose to link up to form a single agent. Note that there is currently significantly limited work in the area of recombinant augmentation, thus making it an open question for future research in the area.

\subsection{Actions}

\begin{figure}
\centering
\begin{subfigure}{.5\textwidth}
  \centering
  \includegraphics[width=0.7\linewidth]{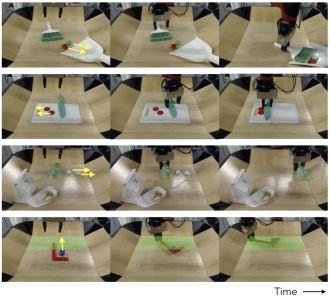}
  \label{fig:sub1}
\end{subfigure}%
\begin{subfigure}{.5\textwidth}
  \centering
  \includegraphics[width=\linewidth]{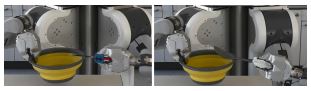}
  \label{fig:sub2}
\end{subfigure}
\caption{Examples of action modification in CPS: Action babbling through repeated interactions with the tool and object \cite{xie2019improvisation} (left), and action modification that adapts known tool use trajectories to new target objects, e.g., from a knife to a spatula \cite{gajewski2019adapting} (right).}
\label{fig:learning}
\end{figure}

In some cases of CPS, the agent may modify the conceptual space associated with actions, i.e., discovering new actions that enables the agent to accomplish its goal. Thus, in these cases the agent \textit{models and reasons about actions}. As opposed to learning from human demonstration or instruction, we review methods for autonomous, fully and/or partially unassisted action discovery. In autonomous action learning, it is important that the agent learns a representation of the task on a motor/trajectory level for its execution, \emph{and} on a semantic level for enabling the agent to determine when the action should be used. Overarching methods of action discovery include action decomposition, and action modification. 

\subsubsection{Action Decomposition} 
In action decomposition, the agent evaluates known or demonstrated actions, to extract useful and reusable substructures. These substructure element(s), in turn, are utilized as standalone actions. CPS methods for decomposition include segmentation, and sub-tasking. 

\smallsec{Segmentation} In the segmentation method of action decomposition, actions are broken into smaller sub-actions through an evaluation of the action's trajectories to identify change points. While several segmentation techniques have been proposed in this context, specifically CPS approaches have used state-based transition autoregressive hidden markov model (STARHMM) \cite{kroemer2015towards} and Bayesian Change-point Detection (BCP) \cite{gizzi2019creative}. In Kroemer et al. \cite{kroemer2015towards}, following segmentation using STARHMM, a combination of RL and Dynamic Movement Primitives (DMPs) are used to sequence new actions and execute them in problem solving.  Gizzi et al.  \cite{gizzi2019creative}, use BCP to segment recorded trajectories of known actions to generate the controllers of candidate action primitives. They ground each controller by evaluating their end effects at a logical descriptor level (using PDDL, see section \ref{subsec:sym}), and consider only those actions which symbolically change the environment to be new action primitives. For example, the action of pushing a button is segmented into a `press' and `release' action, which can then be used in a generalized manner on any object in the environment.

\smallsec{Sub-Tasking} In sub-tasking methods for action decomposition, a high-level task (sequence of actions) is broken down into sub-tasks (smaller sub-sequences of actions). In contrast to action segmentation which deals with individual actions, sub-tasking methods decompose larger tasks into smaller sub-groups of actions (sub-tasks). Xu et al.\cite{xu2018neural}, use neural task programming (NTP) for learning hierarchical decomposition of a task into sub-tasks, which are then generalized to other novel tasks with different task length, topology, and semantics. They demonstrate their method for successful completion of an object sorting, block stacking, and table clean-up task, all of which were initially unseen tasks. In similar work, Hangl et al. \cite{hangl2017skill} first break down larger tasks into smaller sub-behaviors, and later ``compose'' the sub-behaviors to enable a robot to learn new behaviors. Given a set of behaviors $B$, they define new behaviors as a composition of behaviors $b_i \in B$, as $b_l \circ ... \circ b_2 \circ b_1 \circ b^\sigma$. Compound behaviors that are successful at accomplishing the task are added to list of known behaviors for the robot.

\subsubsection{Action Modification} 
Action modification techniques \textit{adapt} pre-existing capabilities to discover novel actions. Modifications can happen at a high level representational level (i.e. operator learning), at a low-level trajectory level, or both. Reviewed methods include behavior babbling adaptations and trajectory/geometry adaptation.

\smallsec{Babbling-based Adaptation}
A common method for autonomous action discovery is `babbling,' where the agent interacts with its environment with the goal of learning about the environment. Action discovery happens through such interactions, in conjunction with validation of new behaviors. Behavior validation can happen through the use of perceptual information, or through intrinsic motivation. In the case of perceptual validation, Gizzi et al. \cite{gizzi2021toward} demonstrate babbling through systematic variations in continuous parameters of known actions, in order to discover new actions to be used in CPS tasks. New actions are validated through novelty/usefulness in predicate end effects. For example, when a ``push'' action results in an object falling off of a surface, the speed parameter can be varied such that a ``nudge'' action is discovered. In the case of tool use in sequential manipulation tasks, predictive models of consequences of actions are developed through repeated interactions of using tools on objects in the environment  \cite{toussaint2018differentiable,allen2019tools,sinapov2007learning,xie2019improvisation}. The known actions are then adapted based on the outcomes of the interactions. The work of Silver et al. \cite{silver2021learning} show how action operators can be learned from data in a task and motion planning (TAMP) domain. Their approach is to learn symbolic actions as probabilistic transition operators, and to learn their controllers through a combination of clustering, predicate search, and parameter estimation. Chitnis et al. \cite{chitnis2021glib} use a technique called ``goal literal babbling'' (GLIB) to learn object-relational transition models to enable generalizable (lifted) planning. Intrinsically motivated goal-based exploration is driven by a novelty measure, encouraging babbling in unvisited state space as a way to learn a complete model for general action policies. The agent is able to use GLIB to update its previously flawed transition model in order to handle novelty, where formerly learned actions are used on novel objects. Qin et al. \cite{qin2020keto} learn novel ways to use tools through repeated, self-supervised interactions with the tools in a simulated environment. Oddi et al. \cite{oddi2020integrating} develop a framework for learning skills through intrinsically motivated reinforcement learning (via self-generated goals), and then abstracting those newly acquired skills to high level, set theoretic action representations. They demonstrate CPS in a grasp task, where the robot is able to learn how to pick up a newly encountered object (vase shaped rock) which exceeds is gripper span. The newly acquired action is encoded in PDDL for generalized use. 

\smallsec{Trajectory/Geometric Adaptation}
In some cases of action modification, the agent adapts known symbolic/non-symbolic action representations to discover new actions. Prior work by Suarez-Hernandes \cite{suarez2020strips} discover new STRIPS actions from execution traces, by introducing cost functions to effectively search the action space. The discovery process consists of four phases: ``Initialization'' that involves generating a new planning problem; ``Searching'' for new actions and validating whether they satisfy the planning problem; ``Expanding'' which involves adding new compilations to an existing open list; and ``Induction'' where the finalized set of actions are induced based on the best planning solution. Prior work has also looked at adapting non-symbolic representations of tool-use trajectories to new tool use scenarios, either adapting to new tool specifications such as a different handle size \cite{fitzgerald2019human,fitzgerald2017human} or to a completely new tool \cite{gajewski2019adapting}.


\renewcommand{\tabcolsep}{3pt}
\begin{table}[]
\centering
\begin{tabular}{@{}l|l|l|l|l|l|@{}}
\cmidrule(l){2-6}
\textbf{}                       & \textbf{Formulation} & \textbf{Representation} & \textbf{Manipulation} & \textbf{Evaluation} & \textbf{Concepts} \\ \midrule
\multicolumn{1}{|l|}{CreaCogs}  & N/a                  & Symbolic                & Combinational         & N/a                 & Objects    \\ \midrule
\multicolumn{1}{|l|}{Robogyver} & Planning         & Non-symbolic            & Combinational         & Physical          & Objects    \\ \midrule
\end{tabular}
\caption{Table highlighting the classifications of the frameworks discussed in Section \ref{sec:examples}}
\end{table}

\section{Examples of Existing Creative Problem Solving Architectures}
\label{sec:examples}
In this section, we describe two existing architectures specifically targeted for creative problem solving, namely CreaCogs and Robogyver. While there exist other architectures such as ICARUS and DIARC that can support creative reasoning, we do not include them here since they are not focused specifically on CPS but rather general planning.

\subsection{CreaCogs}
\label{subsec:Creacogs}
CreaCogs is an architecture that enables agents to solve creative problems through Object Replacement and Object Construction (OROC) \cite{oltecteanu2016object}. CreaCogs seeks to accomplish three creative tasks: i) replacement of a missing object for a task; ii) composing objects to construct a final object for a task; and iii) decomposing objects. The architecture reasons about object affordances and capabilities in order to accomplish each task. CreaCogs consists of a symbolic large-scale knowledge base that organizes concepts at three levels: i) a feature space that encodes features or attributes of objects; ii) a concept level that represents all the relationships between the known features of an object; and iii) a problem template where the relations between different concepts are encoded. In their work, the information regarding the three levels is symbolically encoded a-priori. Given the knowledge base, replacement objects are identified on the basis of the similarity of their features to the missing object. Similarly, composite objects are represented as a conjunction of the features (denoted as $Y$) of the individual objects (denoted as $O_k$) that make up the composite object, e.g., \textit{relation($O_1$, Y) $\land$ relation($O_2$, Y)}. Hence, CreaCogs uses combinational manipulation for identifying composite objects. Additionally, topological relationships are used to define the relative positioning of the objects when constructing the composite object. Lastly, for decomposition, a larger object is represented as a conjunction of its individual components, e.g., \textit{object(fishing rod)} $\rightarrow$ \textit{component(hook, fishing rod)} $\land$ \textit{component(line, fishing rod)} $\land$ \textit{component(rod, fishing rod)} etc. Thus, new concepts (i.e., objects) can be identified through various combinations of relevant concepts. 


In Olte{\c{t}}eanu \cite{olteteanu2015seeing}, the authors discuss re-representation of the problem in order to highlight concepts that are relevant for object replacement and composition, e.g., ``containability'' is a relevant concept when creating a candle holder. For this problem, a candle holder is re-represented to highlight the containability feature that then helps identify substitute objects for it. They discuss the notion of ``seeing as'' referring to: i) the ability to represent a group of features as a meaningful object, and ii) ability to represent a group of objects as a meaningful structure that can solve the problem on hand. Thus, the initial problem description is \textit{transformed} into an alternate representation that highlights the relevant concepts or features of object(s) that can enable the agent to solve the problem at hand. While the authors discuss these ideas theoretically in \cite{olteteanu2015seeing}, the CreaCogs implementation in \cite{oltecteanu2016object} does not include the re-representation, hence we categorize it as combinational manipulation within our survey.

\subsection{Robogyver}
\label{subsec:Robogyver}


Robogyver is an architecture that enables robots to accomplish tasks where required tools are missing, either through tool construction or tool substitution \cite{nair2020tool,nair2021defense}. While substitution and construction is closely related to object replacement and composition in CreaCogs, Robogyver focuses specifically on tool-based problems through non-symbolic rather than symbolic reasoning \cite{nair_FGS}. Robogyver formulates tool macgyvering as a planning problem \cite{nair_FGS}, introducing the Feature Guided Search (FGS) approach that extends classical planning using supervised learning to guide the search through the planning space. The initial conceptual space comprises of multi-modal sensory inputs (object point clouds and spectral readings). The authors perform multi-objective optimization in order to identify viable objects for replacing the missing tool, based on their point clouds (indicating shape) and spectral measurements (indicating material). The objectives within the multi-objective function are learned through supervised learning techniques. Robogyver uses a hybrid symbolic and non-symbolic representation wherein the objects are represented in a non-parametric and non-symbolic manner, whereas, the task planning itself uses symbols represented in PDDL. To perform tool construction, the multi-objective functions are used to evaluate object combinations to identify viable constructions. Specifically, the multi-objective function, denoted as $f$, acts as a combinational function that takes object point clouds and spectral readings as inputs, i.e., $f(o_1, o_2, ..., o_m)$ (i.e., combinational methods). Within the new conceptual space, the multi-objective functions are used to guide the search for a valid task plan. Specifically, the object-based symbols within the planner are assigned the output scores of the multi-objective functions. The scores are then incorporated into planning heuristics in order to guide the search (using $A^*$) towards valid object combinations. Hence, given a task and a set of objects, the agent is able to adapt the task plan to construct an appropriate tool for the task. The approach is evaluated on a physical robot (Kinova) for the construction of tools for six different tasks.

\section{Open Research Questions}
\label{sec:future}
In our survey of creative problem solving, we identified open research questions that have not been adequately addressed in the field of CPS. In highlighting these problems, we hope to stimulate further work and suggest future research directions. We divide this section into two subsections. The first subsection discusses open research questions in the context of our CPS framework. The second subsection discusses broader CPS questions that are not specifically related to our framework or taxonomy, but need to be addressed, and potentially incorporated into future iterations of the CPS framework.

\subsection{Open Questions Relating to CPS Framework}

\begin{figure}[t]
	\centering
	\includegraphics[width=0.95\textwidth]{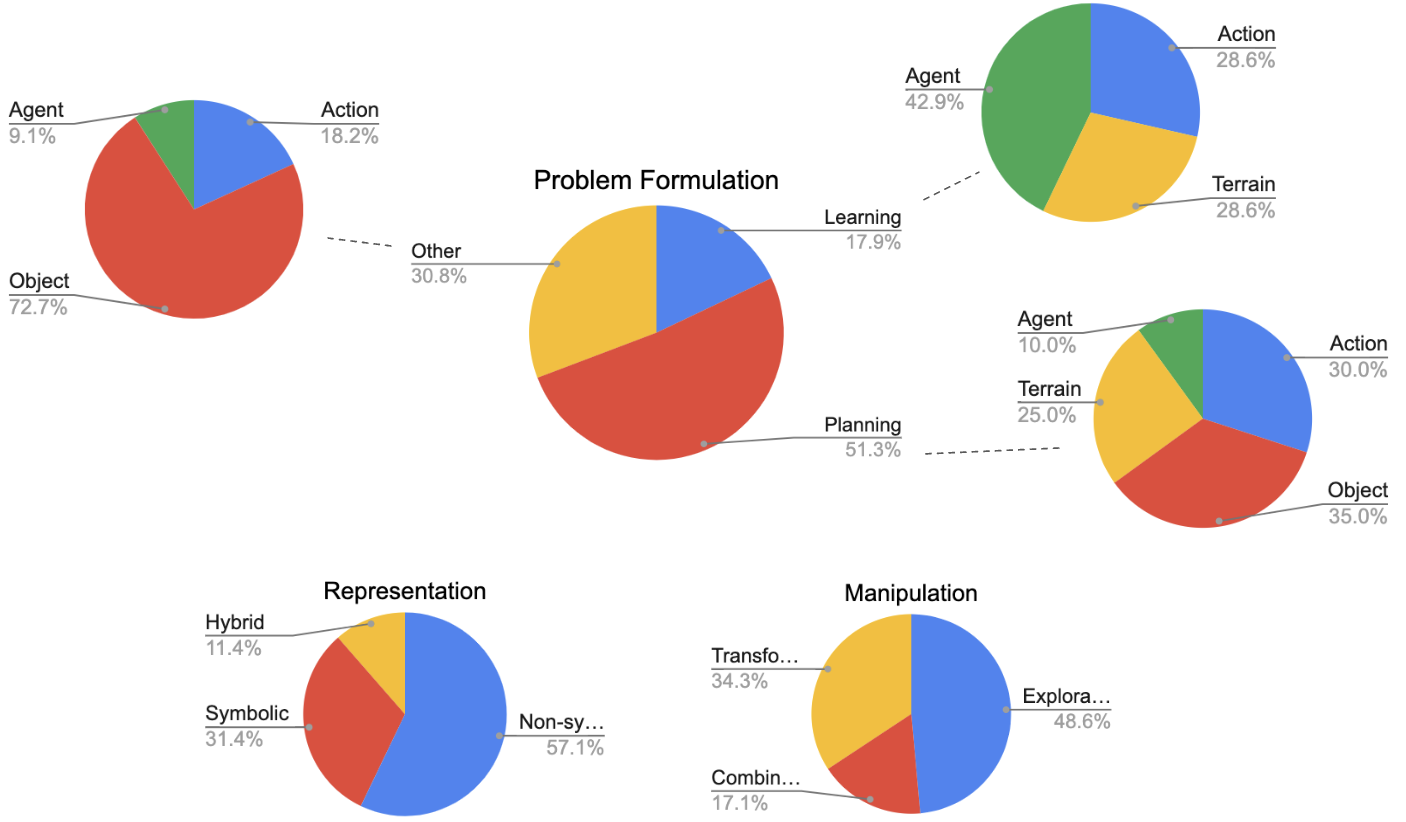}
	\captionsetup{width=\linewidth}
	\caption{Distribution of the 51 papers in our survey. To note limited work in CPS, other paradigms of knowledge manipulation have not focused on terrain modification, and learning approaches have not looked at object modification indicating clear gaps in existing CPS research. A tabulated list of papers in each category is provided in the Appendix.}
	\label{fig:piechart}
\end{figure}

In this section, we discuss open research questions in the context of our CPS taxonomy.

\subsubsection{Hybrid Symbolic and Non-symbolic Representations}
There has been limited work utilizing hybrid symbolic and non-symbolic representations in CPS. Such hybrid representations can be useful in order to effectively represent different types of information required by the CPS method. In particular, some types of information such as visual properties are more easily represented through non-symbolic means, whereas symbolic representations can be useful to encode hierarchies such as in semantic networks. 

There are several challenges to further explore in this space. In particular, what specific information should be represented as symbolic vs. non-symbolic (or vice-versa)? Moreover, how can we design hybrid systems that can effectively leverage the strengths of the hybrid representation? As noted in our survey, planning approaches are strongly correlated to symbolic representations, whereas learning approaches are correlated to non-symbolic representations. Hence, hybrid systems that combine planning and learning, can be useful (and perhaps necessary) for effectively utilizing hybrid representations. To this end, existing approaches such as hybrid planning and RL \cite{strens2004combining} can be beneficial. Additionally, recent advances in Neuro-symbolic AI \cite{d2020neurosymbolic}, can also be useful in this context. Neuro-symbolic AI seeks to combine neural networks with symbolic representations of the problem, and can greatly enhance the reasoning capability, as well as explanability of CPS systems.

\subsubsection{Rethinking Agent-Reasoning in CPS}
Among the conceptual spaces discussed in our survey, the papers relating to conceptual spaces of agent states, focus on agent modification alone. However, a key point to note is that modeling and reasoning about the agent can have wider implications for CPS, beyond just agent modification. Modeling the capabilities of the agent/robot, can be highly beneficial for identifying practically viable creative solutions for performing the task, which is especially important when operating in real-world domains rather than simulations. A key question here is, how can CPS systems account for the agent when deriving solutions? In this case, the solutions derived would depend on the agent's capabilities, e.g., it could vary from the two-armed Baxter robot to a single-armed Sawyer robot. Consider the example of opening a tightly closed jar. The Baxter robot can potentially grasp the jar with one hand and use a knife with the other to pry it open. In contrast, this would not be a viable solution for Sawyer, that may instead run it under hot water to loosen the cap, or break it open to access its contents. However, existing CPS research does not account for the agent's capabilities in this manner, making it an open question for future work.

\subsection{Broader CPS Research Questions}
In this section, we discuss broader CPS questions relating to, a) universal metric for evaluating CPS solutions, b) lifelong CPS, and c) generalizability in CPS.

\subsubsection{Universal Metric for CPS Solutions}
As described in Section \ref{sec:theory}, there does not yet exist a universally agreed upon definition of computational creativity. Because of this lack of a concise definition of CC, it has been challenging for the CPS research community to develop a universal measure of a CPS solution, with consideration of \emph{both} the traditional problem solving aspects of CPS, and the creativity portion of CPS. Most evaluation metrics of CPS which have been developed in past research are domain and/or problem specific (See Section \ref{subsec:benchmark}). While some degree of universality has been considered, most work has focused on developing tests for CPS success, rather than a specific metric for objectively quantifying how \emph{creative} a solution is. Other works have developed \emph{theoretical} measures of CPS solutions, without evidentiary implementation. Developing such a metric for CPS would be highly beneficial, as it would help streamline and benchmark existing and future CPS methods. For instance, an objective metric for categorizing the subset of papers in RL that constitutes CPS needs to be defined, as opposed to the subjective novelty-based criterion used in this survey. 

\subsubsection{Lifelong CPS}
Techniques in CPS have focused on flexibility in handling unforeseen and ill-structured problems. Despite extensive research into creating flexible CPS methods for ``problem at hand'' solutions, there has not been extensive research into methods for using past CPS encounters to improve future CPS performance, thus employing ``lifelong'' learning for CPS. For example, many CPS methods surveyed involve environment exploration, used for optimizing a specific task solution. An open question here is, how can the agent learn from, and adapt its prior experiences to effectively solve CPS tasks in the future? The agent's prior interactions could be used to support lifelong CPS, by minimizing the amount of environment interaction that the agent may need in a different, future CPS task. Existing research in lifelong learning \cite{parisi2019continual} and transfer learning \cite{transfer_survey} could be useful venues to explore in this context. Future research should consider how agents can improve their own CPS abilities as they are continually put into scenarios where they need to discover new information. 

\subsubsection{Generalizability and CPS}
A key consideration and theme in CPS research is the flexibility of systems, which allows agents to handle the inherent novelty of CPS tasks. While current research has made this consideration of generalizabilty within individual problem domains, there has yet to be extensive research and testing of cross-domain generalizablity of CPS systems. Additionally, despite the domain-agnostic theoretical formulations of CPS methods which exist in current research, testing specific methods for cross-domain generalizablity at the implementation level remains a largely unexplored challenge.


\section{Conclusions}
This survey discussed creative problem solving in artificial intelligence, by formally defining CPS, and introducing a framework that encompasses existing CPS approaches. In contrast to existing CPS formalizations, our definition 1) introduces a broader notion of concepts, and 2) connects Boden's three levels of creativity from computational creativity literature, to problem solving in AI, thus leveraging theoretical aspects from both CC and AI. Our presented CPS framework consists of four key steps including, 1) problem formulation, 2) knowledge representation, 3) knowledge manipulation, and 4) evaluation. Additionally, we categorized existing CPS research within our framework. We further expanded on our CPS framework to discuss the two types of conceptual spaces that are modified, i.e., states (including objects, terrain and agent) and actions. We further categorized existing approaches along this dimension. Next, we presented existing CPS architectures, also organizing them within our CPS taxonomy. Our survey of the literature concluded with a list of open research questions, which we believe will serve as a useful guide for future work in CPS. We further hope that this survey will encourage research into this relatively unexplored field, bridging the gap between CC and problem solving in AI.

\begin{appendices}

\renewcommand{\tabcolsep}{3pt}
\begin{longtable}{@{}l|l|l|l|l|l|@{}}
\midrule
\multicolumn{1}{|l|}{\textbf{Paper}}        & \textbf{Formuln.} & \textbf{Repn.} & \textbf{Manipn.} & \textbf{Eval.} & \textbf{Concepts} \\ \midrule
\multicolumn{1}{|l|}{\cite{xie2019improvisation}}    & Learning             & Non-symbolic                & Expl           & Phys          & Action   \\ \midrule
\multicolumn{1}{|l|}{\cite{fitzgerald2014representing,fitzgerald2017human}}    & Planning             & Non-symbolic                & Trans           & Phys          & Action   \\ \midrule
\multicolumn{1}{|l|}{\cite{wicaksono2017towards}}    & Planning             & Symbolic                & Expl           & Phys          & Object   \\ \midrule
\multicolumn{1}{|l|}{\cite{wicaksono2020cognitive}}    & Planning             & Symbolic                & Expl           & Phys          & Object   \\ \midrule
\multicolumn{1}{|l|}{\cite{saboia2019autonomous}}    & Planning             & Non-symbolic                & Trans           & Phys          & Terrain   \\ \midrule
\multicolumn{1}{|l|}{\cite{toussaint2018differentiable}}    & Planning             & Hybrid                & Expl           & Sim          & Action   \\ \midrule
\multicolumn{1}{|l|}{\cite{zhu2015understanding}}    & Other             & Hybrid                & Trans           & Sim          & Object   \\ \midrule
\multicolumn{1}{|l|}{\cite{boteanu2015towards}}    & Planning             & Symbolic                & Expl           & Phys          & Object   \\ \midrule
\multicolumn{1}{|l|}{\cite{oltecteanu2016object}}    & Other             & Symbolic                & Comb.           & n/a          & Object   \\ \midrule
\multicolumn{1}{|l|}{\cite{erdogan2013planning}}    & Planning             & Symbolic                & Expl           & Sim          & Terrain   \\ \midrule
\multicolumn{1}{|l|}{\cite{erdogan2016autonomous}}    & Planning             & Symbolic                & Expl           & Sim          & Object   \\ \midrule
\multicolumn{1}{|l|}{\cite{choi2018creating}}    & Planning             & Non-symbolic                & Expl           & Sim          & Terrain   \\ \midrule
\multicolumn{1}{|l|}{\cite{levihn2015using}}    & Planning             & Symbolic                & Expl           & Sim          & Object   \\ \midrule
\multicolumn{1}{|l|}{\cite{tosun2018perception}}    & Planning             & Non-symbolic                & Trans           & Phys          & Terrain   \\ \midrule
\multicolumn{1}{|l|}{\cite{abelha2016model}}    & Other             & non-symbolic                & Trans           & Sim          & Object   \\ \midrule
\multicolumn{1}{|l|}{\cite{schoeler2015bootstrapping}}    & Other             & Non-symbolic                & Trans           & Sim          & Object   \\ \midrule
\multicolumn{1}{|l|}{\cite{gajewski2019adapting}}    & Planning             & Non-symbolic                & Trans           & Phys          & Action   \\ \midrule
\multicolumn{1}{|l|}{\cite{baker2019emergent}}    & Learning             & Non-symbolic                & Expl           & Sim          & Terrain   \\ \midrule
\multicolumn{1}{|l|}{\cite{allen2019tools}}    & Learning             & Non-symbolic                & Expl           & Sim          & Action   \\ \midrule
\multicolumn{1}{|l|}{\cite{sinapov2007learning}}    & Other             & Non-symbolic                & Trans           & Sim          & Action   \\ \midrule
\multicolumn{1}{|l|}{\cite{hangl2017skill}}    &     Other         & Non-symbolic                & Comb.           & Phys          & Action   \\ \midrule
\multicolumn{1}{|l|}{\cite{bapst2019structured}}    & Learning             & Non-symbolic                & Expl           & Sim          & Terrain   \\ \midrule
\multicolumn{1}{|l|}{\cite{pathak2019learning}}    & Learning             & Non-symbolic                & Expl           & Sim          & Agent   \\ \midrule
\multicolumn{1}{|l|}{\cite{nair2019tool,nair2019autonomous,nair2020tool}}    & Other             & Symbolic                & Comb.           & Phys          & Object   \\ \midrule
\multicolumn{1}{|l|}{\cite{nair_FGS}}    & Planning             & Hybrid                & Comb.           & Phys          & Object   \\ \midrule
\multicolumn{1}{|l|}{\cite{shrivatsav2019tool}}    & Other             & Non-symbolic                & Comb.           & Phys          & Object   \\ \midrule
\multicolumn{1}{|l|}{\cite{gizzi2019creative}}    & Planning             & Hybrid                & Trans           & Sim          & Action   \\ \midrule
\multicolumn{1}{|l|}{\cite{gizzi2021toward}}    & Planning             & Hybrid                & Expl           & Sim          & Action   \\ \midrule
\multicolumn{1}{|l|}{\cite{murookaself}}    & Planning             & Non-symbolic                & Trans           & Phys          & Agent   \\ \midrule
\multicolumn{1}{|l|}{\cite{suarez2020strips}}    & Planning             & Symbolic                & Expl           & Sim          & Action   \\ \midrule
\multicolumn{1}{|l|}{\cite{freedman2020}}    & Planning             & Symbolic                & Trans           & Sim          & Object  \\ \midrule
\multicolumn{1}{|l|}{\cite{ha2019reinforcement}}    & Learning             & Non-symbolic                & Expl           & Sim          & Agent   \\ \midrule
\multicolumn{1}{|l|}{\cite{zhao2020robogrammar}}    & Other             & Symbolic                & Expl           & Sim          & Agent   \\ \midrule
\multicolumn{1}{|l|}{\cite{yang2020autonomous}}    & Other             & Non-symbolic                & Comb.           & Phys          & Object   \\ \midrule
\multicolumn{1}{|l|}{\cite{qin2020keto}}    & Planning             & Non-symbolic                & Trans           & Sim          & Action   \\ \midrule
\multicolumn{1}{|l|}{(Colin et al., 2016)}    & Learning             & Non-symbolic                & Comb.           & Sim          & Terrain   \\ \midrule
\multicolumn{1}{|l|}{\cite{colin2019reinforcement}}    & Learning             & Non-symbolic                & Comb.           & Sim          & Terrain   \\ \midrule
\multicolumn{1}{|l|}{\cite{lieto2019beyond}}    & Planning             & Symbolic                & Comb.           & B.mark          & Object   \\ \midrule
\multicolumn{1}{|l|}{\cite{silver2021learning}}    & Planning             & Hybrid                & Trans           & Sim          & Action   \\ \midrule
\multicolumn{1}{|l|}{\cite{chitnis2021glib}}    & Learning             & Symbolic                & Expl           & Sim          & Action   \\ \midrule
\multicolumn{1}{|l|}{\cite{oddi2020integrating}}    & Learning             & Hybrid                & Expl           & Sim          & Action   \\ \midrule
\multicolumn{1}{|l|}{\cite{kralik2016modeling}}    & Learning             & Non-symbolic                & Trans           & B.mark          & Terrain   \\ \midrule
\multicolumn{1}{|l|}{\cite{xu2018neural}}    & Other             & Hybrid                & Trans           & Sim          & Object   \\ \midrule
\multicolumn{1}{|l|}{\cite{fitzgerald2019human}}    & Learning             & Non-symbolic                & Trans           & Phys          & Action   \\ \midrule
\multicolumn{1}{|l|}{\cite{kroemer2015towards}}    & Learning             & Non-symbolic                & Trans           & Phys          & Action   \\ \midrule
\multicolumn{1}{|l|}{\cite{ha2019reinforcement}}    & Learning             & Non-symbolic                & Expl           & Sim          & Agent   \\ \midrule
\multicolumn{1}{|l|}{\cite{schaff2019jointly}}    & Learning             & Non-symbolic                & Expl           & Sim          & Agent   \\ \bottomrule
\end{longtable}

\end{appendices}

\bibliography{orig}
\bibliographystyle{theapa}

\end{document}